\documentclass[10pt,twocolumn,letterpaper]{article}

\usepackage{cvpr}

\usepackage{times}
\usepackage{relsize}
\usepackage[T1]{fontenc}
\usepackage[latin1]{inputenc}
\usepackage[english]{babel}

\usepackage{epsfig}
\usepackage{graphicx}
\usepackage{wrapfig}
\usepackage[belowskip=1pt,aboveskip=5pt,font=small]{caption}
\usepackage[belowskip=0pt,aboveskip=3pt,font=small]{subcaption}
\usepackage{amsmath, amsthm, amssymb, mathabx}
\usepackage{textcomp}
\usepackage{stmaryrd}
\SetSymbolFont{stmry}{bold}{U}{stmry}{m}{n}
\usepackage{upgreek}
\usepackage{bm}
\usepackage{cases}
\usepackage{mathtools}
\usepackage{arydshln}
\usepackage{multirow}

\usepackage{cite}
\usepackage[pagebackref=true,breaklinks=true,letterpaper=true,colorlinks,bookmarks=false,citecolor=green]{hyperref}
\usepackage{bibspacing}
\usepackage{algorithm}
\usepackage{algpseudocode}

\usepackage{multirow}
\usepackage{rotating}
\usepackage{booktabs}

\usepackage{enumitem}
\usepackage[olditem,oldenum]{paralist}

\usepackage{alltt}
\usepackage{listings}

\usepackage{mysymbols}

\usepackage{url}
\usepackage{xspace}
\usepackage{comment}
\usepackage{color}
\usepackage[dvipsnames]{xcolor}
\usepackage{afterpage}
\usepackage{pdfpages}
\usepackage{framed}
\usepackage{fancybox}
\usepackage{cuted}
\usepackage{caption}

\usepackage{quoting}
\quotingsetup{vskip=0pt}
\usepackage{epigraph}

\setdefaultleftmargin{.2cm}{.2cm}{}{}{}{}
\setlist{leftmargin=.2cm}

\newcommand{\vdfull}{Visual \Dialog}
\newcommand{\vqa}{VQA\xspace}

\newcommand{\myquote}[1]{\emph{`#1'}}
\newcommand{\myapprox}{{\raise.17ex\hbox{$\scriptstyle\sim$}}}
\newcommand{\xhdr}[1]{\vspace{3pt}\noindent\textbf{#1}}

\newcommand{\reffig}[1]{Fig.~\ref{#1}}
\newcommand{\refsec}[1]{Sec.~\ref{#1}}

\newcommand{\eqa}{EmbodiedQA\xspace}

\newcommand{\eqads}{EQA\xspace}

\belowdisplayskip \abovedisplayskip

\newlength{\sectionReduceTop}
\newlength{\sectionReduceBot}
\newlength{\subsectionReduceTop}
\newlength{\subsectionReduceBot}
\newlength{\abstractReduceTop}
\newlength{\abstractReduceBot}
\newlength{\captionReduceTop}
\newlength{\captionReduceBot}
\newlength{\subsubsectionReduceTop}
\newlength{\subsubsectionReduceBot}

\newlength{\eqnReduceTop}
\newlength{\eqnReduceBot}

\newlength{\horSkip}
\newlength{\verSkip}

\newlength{\figureHeight}
\newcommand{\ad}[1]{\textcolor{black}{#1}}
\newcommand{\sml}[1]{\textcolor{black}{#1}}

\newcommand{\devi}[1]{\textcolor{black}{#1}}
\newcommand{\FIXME}[1]{\textbf{\textcolor{black}{FIX: #1}}}
\newcommand{\verify}[1]{{\textcolor{black}{#1}}}

\cvprfinalcopy 

\def\cvprPaperID{**} 

\ifcvprfinal\fi
\usepackage{flushend}

\begin{document}

\title{Embodied Question Answering}

\author{
    Abhishek Das$^{1^\star}$, \,\,
    Samyak Datta$^1$, \,\,
    Georgia Gkioxari$^2$, \,\,
    Stefan Lee$^1$, \,\,
    Devi Parikh$^{2,1}$, \,\,
    Dhruv Batra$^{2,1}$ \vspace{3pt}\\
    $^1$Georgia Institute of Technology, $^2$Facebook AI Research\\
    {\tt\small $^1$\{abhshkdz, samyak, steflee\}@gatech.edu}
    \quad \tt\small $^2$\{gkioxari, parikh, dbatra\}@fb.com \\
    \tt\normalsize \href{https://embodiedqa.org}{embodiedqa.org}
}

\maketitle
\renewcommand*{\thefootnote}{$\star$}
\setcounter{footnote}{1}
\footnotetext{Work partially done during an internship at Facebook AI Research.}
\renewcommand*{\thefootnote}{\arabic{footnote}}
\setcounter{footnote}{0}
\thispagestyle{empty}

\begin{abstract}

We present a new AI task -- \textbf{Embodied Question Answering} (\eqa) --
where an agent is spawned at a random location in a 3D environment and
asked a question (\myquote{What color is the car?}).
In order to answer, the agent
must first intelligently navigate to explore the environment,
gather information through first-person (egocentric) vision, and then answer the
question (\myquote{orange}).

This challenging task requires a range of AI skills --
active perception, language understanding, goal-driven navigation,
commonsense reasoning, and grounding of language into actions.
In this work, we develop the environments, end-to-end-trained reinforcement learning agents, and evaluation protocols for \eqa.


\end{abstract}

\section{Introduction}
\label{sec:intro}

%

\setlength{\epigraphwidth}{0.85\columnwidth}
\renewcommand{\textflush}{flushepinormal}
\epigraph{The embodiment hypothesis is the idea that intelligence emerges in the interaction of an agent with
an environment
and as a result of sensorimotor activity.}
{\textit{Smith and Gasser~\cite{smith_al05}}}

\vspace{-8pt}
%

Our long-term goal is to build intelligent agents
that can \emph{perceive} their environment (through vision, audition, or other sensors),
\emph{communicate} (\ie, hold a natural language dialog grounded in the environment),
and \emph{act} (\eg aid humans by executing API calls or commands in a virtual or embodied environment).
In addition to being a fundamental scientific goal in artificial intelligence (AI),
even a small advance towards such intelligent systems can \emph{fundamentally change our lives} --
from assistive dialog agents for the visually impaired, to natural-language interaction with self-driving cars,
in-home robots, and personal assistants.

As a step towards goal-driven agents that can perceive, communicate, and
execute actions, 
we present a new AI task -- \textbf{\emph{Embodied Question Answering}} (\eqa), along with virtual
environments, evaluation metrics, and a novel deep reinforcement learning (RL) model for this task.

\begin{figure}[t]
    \centering
    \includegraphics[width=\columnwidth]{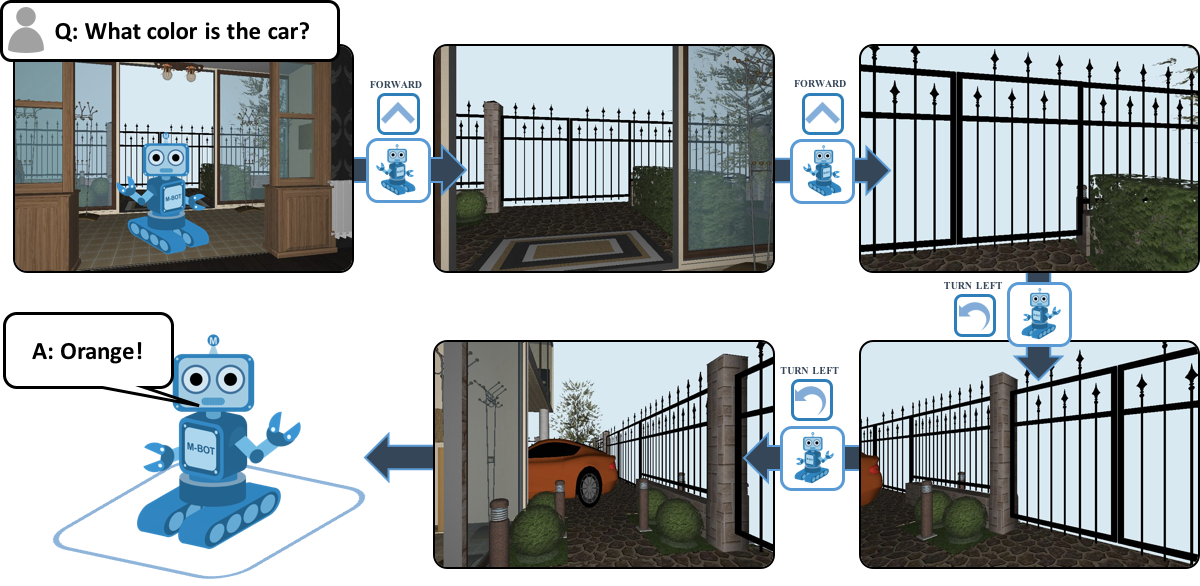}\\[10pt]
    \caption{Embodied Question Answering -- \eqa -- tasks agents with navigating rich 3D environments
    in order to answer questions. These embodied agents must jointly learn language understanding,
    visual reasoning, and navigation to succeed.}
    \label{fig:teaser}
\end{figure}

Concretely, the \eqa task is illustrated in \figref{fig:teaser} -- an agent is spawned at a random location
in an environment (a house or building)
and asked a question (\eg \myquote{What color is the car?}).
The agent perceives its environment through first-person vision (a single RGB camera)
and can perform a few atomic actions:
move-$\{$forward, backward, right, left$\}$ and turn-$\{$right, left$\}$.
The goal of the agent is to intelligently navigate the environment and gather
the visual information necessary 
to answer the question.


\eqa is a challenging task that subsumes several fundamental AI problems as sub-tasks.
Clearly, the agent must understand language (\emph{what is the question asking?})
and vision (\emph{what does a car look like?}), but a successful agent must also learn to perform:\\[-20pt]

\begin{changemargin}{0.125cm}{0.125cm}
\textbf{Active Perception}: 
The agent may be spawned anywhere in the environment and may not immediately `see'
the pixels containing the answer to the visual question (\ie the car may not be visible).
Thus, the agent \emph{must} move to succeed -- controlling the pixels that it will perceive.
The agent must learn to map its visual input to the correct action based on its perception of
the world, the underlying physical constraints, and its understanding of the question.

\textbf{Common Sense Reasoning}: The agent is not provided a floor-plan or map of the environment,
and must navigate from egocentric views alone. Thus,
it must learn common sense (\emph{where am I? where are cars typically found in a housing compound?
and where is the garage with respect to me?})
similar to how humans may navigate in a house they have never visited (\emph{the car is probably in the garage outside,
so I should find a door that leads out}).

\textbf{Language Grounding:}
One commonly noted shortcoming of modern vision-and-language 
models
is their lack of grounding -- these models often fail to associate entities in text with corresponding image pixels,
relying instead on dataset biases to respond seemingly intelligently even when attending to irrelevant regions~\cite{das_emnlp16, goyal_cvpr17}.
%
In \eqa, we take a goal-driven view of grounding -- our agent grounds a visual question not into pixels
but into a sequence of actions (`garage' \emph{means} to navigate towards the house exterior where the `car'
is usually parked).

\textbf{Credit Assignment:}
From a reinforcement learning perspective, \eqa presents a particularly
challenging learning problem.
Consider the question \myquote{How many rooms contain chairs?}.
How does an agent discover that this question involves exploring the environment to visit `rooms',
detecting `chairs', incrementing a count every time a `chair' is in the view (except while the agent is in the same `room'),
and stopping when no more `rooms' can be found?
All without knowing what a `room' is or how to find it, what a `chair' looks like, or what counting is.
To succeed, the agent must execute a somewhat precise sequence of hundreds of
inter-dependent actions (forward, forward, turn-right, forward, forward, \ldots, turn-left, `5')
-- all to be learned from a reward signal that says `4' is the right answer and anything else is incorrect.
The task is complex enough that most random action sequences
result in negative reward, and when things do go wrong, it's difficult for the agent to know why --
was the question misunderstood? Can the agent not detect chairs? Did the agent navigate incorrectly? Was the counting
incorrect?

\end{changemargin}
\vspace{-10pt}

%
As the first step in this challenging space, we
judiciously scope out a problem space -- environments, question types, learning paradigm --
that allow us to
augment the sparse RL rewards with imitation learning (showing the agent example trajectories)
and reward shaping~\cite{ng_icml99}  (giving intermediate `getting closer or farther' navigation rewards).
Specifically, our approach follows the recent paradigm from robotics and deep RL~\cite{levine_jmlr16,dipendra_emnlp17} --
that the training environments are assumed to be sufficiently \emph{instrumented} -- \ie, provide access to the agent
location, depth and semantic annotations of the environment,
and allow for computing obstacle-avoiding shortest paths from the agent to any target location.

Crucially, at test time, our agents operate entirely from egocentric RGB vision alone --
no structured representation of the environments, no access to a map,
no explicit localization of the agent or mapping of the environment,
no A* or any other heuristic planning, and no pre-processing or hand-coded knowledge
about the environment or the task of any kind.
The agent in its entirety -- vision, language, navigation, answering modules -- is trained
completely end-to-end -- from raw sensory input (pixels and words)
to goal-driven multi-room indoor navigation to visual question answering!

\textbf{Contributions.}
We make the following contributions:

\begin{compactitem}

\item We propose a new AI task: \eqa,
where an agent spawned in an environment must intelligently navigate from an egocentric view to
gather the necessary information to answer visual questions about its environment. \\[-10pt]

\item
We introduce a novel Adaptive Computation Time~\cite{graves_arxiv16} navigator --
that decomposes navigation into a `planner' that selects actions,
and a `controller' that executes these primitive actions a variable number of times
before returning control to the planner.
When the agent decides it has seen the required visual information to
answer the question, it stops navigating and outputs an answer. \\[-10pt]



\item We initialize our agents via imitation learning and show that
agents can answer questions more accurately after fine-tuning with
reinforcement learning -- that is, when allowed to control their own navigation
\emph{for the express purpose} of answering questions accurately.
Unlike some prior work, we explicitly test and demonstrate \emph{generalization
of our agents to unseen environments}.
\\[-10pt]

\item We evaluate our agents in \emph{House3D}~\cite{house3d},
a rich, interactive environment based on human-designed 3D indoor scenes
from the SUNCG dataset~\cite{song_cvpr17}.
These diverse virtual environments enable us to
test generalization of our agent across floor-plans, objects, and room configurations --
without the concerns of safety, privacy, and expense inherent to real robotic platforms. \\[-10pt]

\item
We introduce the \eqads dataset of
visual questions and answers grounded in House3D. The different question types test a range of agent abilities --
scene recognition (\texttt{location}), spatial reasoning (\texttt{preposition}), color recognition (\texttt{color}).
While the \eqa task definition supports free-fom natural language questions, we represent each
question in \eqads as as a \emph{functional program} that can be programmatically
generated and executed on the environment to determine the answer.  
This gives
us the ability to control the distribution of
question-types and answers in the dataset,
deter algorithms from exploiting dataset bias~\cite{zhang_cvpr16, goyal_cvpr17},
and provide fine-grained breakdown of performance by skill. \\[-10pt]

\item
We integrated House3D renderer with Amazon Mechanical Turk (AMT), allowing subjects to
\emph{remotely operate the agent}, and collected expert demonstrations
of question-based navigation that
serve as a benchmark to compare our proposed and future algorithms.

All our code and data will be made publicly available.

\end{compactitem}

\section{Related Work}
\label{sec:related}

\vspace{6pt}
\begin{minipage}{\columnwidth}
\begin{minipage}{0.45\textwidth}
We place our work in context by arranging prior work along the axes of
vision (from a single-frame to video), language (from single-shot question answering to dialog),
and action (from passive observers to active agents).
When viewed from this perspective, \eqa
\end{minipage}\hspace{0.03\textwidth}%
\begin{minipage}{0.53\textwidth}
\vspace{-2pt}
\includegraphics[width=0.97\textwidth,clip=true, trim=8px 0px 0px 5px]{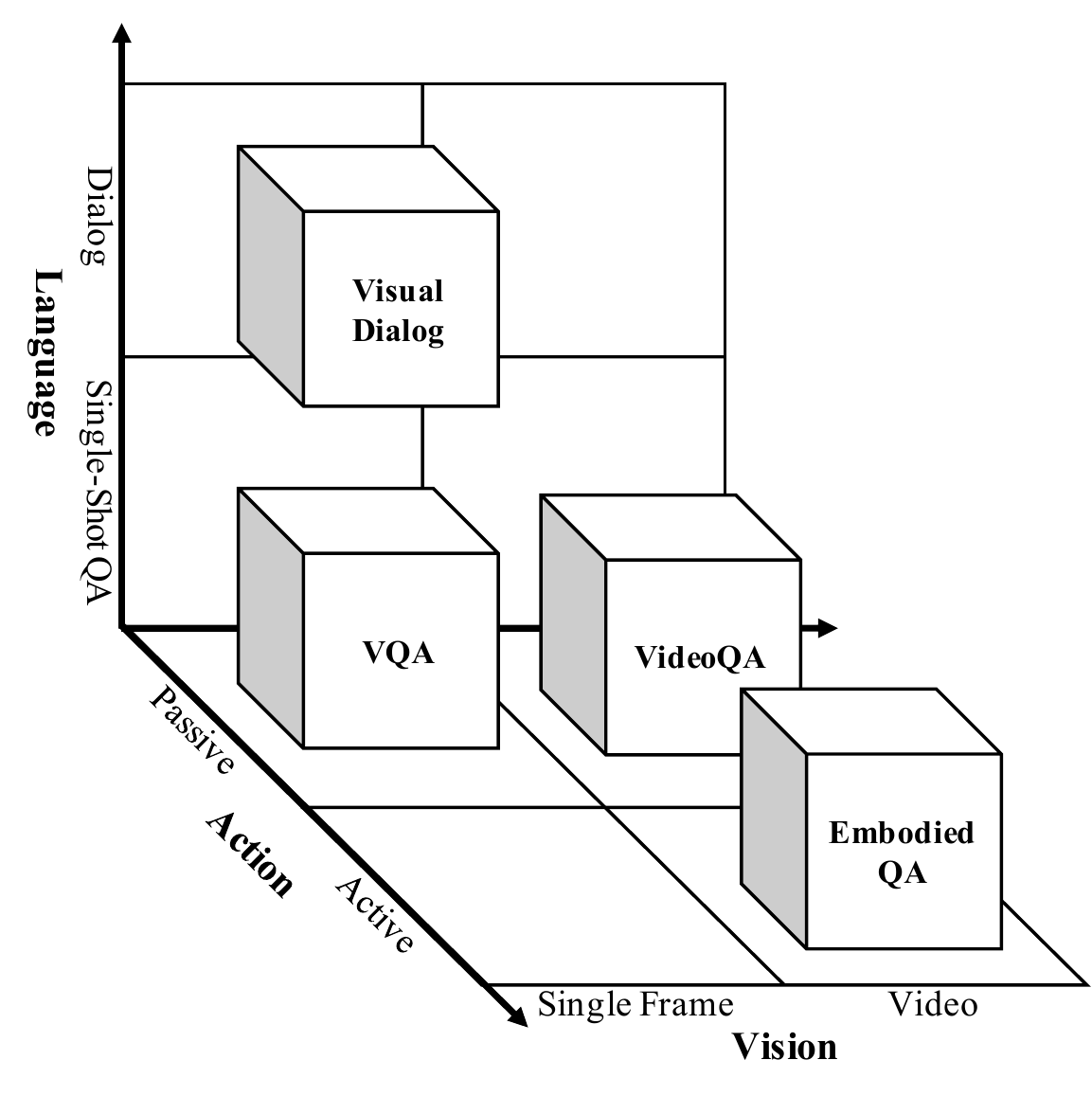}%
\label{fig:related}
\end{minipage}

\end{minipage}\\[2pt]
presents a novel problem configuration -- single-shot QA about videos captured by goal-driven
active agents. Next, we contrast this against various 2D slices in this space.

\xhdr{VQA: Vision + Language.}
Like \eqa, image and video question answering tasks \cite{antol_iccv15, fukui_emnlp16, anderson_arxiv17,
tapaswi_cvpr16, jang_cvpr17} require reasoning about natural language questions
posed about visual content.
The crucial difference is the \emph{lack of control} --
these tasks present answering agents with
a fixed view of the environment (\ie one or more images from some fixed trajectory through
the world) from which the agent must answer the question, never allowing the agents to
\emph{actively} perceive.
In contrast, \eqa agents control their trajectory and fate, for good or ill. The task
is significantly harder than VQA (\ie most random paths are useless) but the agent
has the flexibility to avoid confusing viewpoints and seek visual input that will maximize answer confidence.


\xhdr{Visual Navigation: Vision + Action.}
The problem of navigating in an environment based on visual perception has long been studied in
vision and robotics (see \cite{thrun2005probabilistic} for an extensive survey).
Classical techniques divide navigation into two distinct phases --
mapping (where visual observations are used to construct a 3D model of the environment), and
planning (which selects paths based on this map).
Recent developments in deep RL have proposed fused architectures
that go directly from egocentric visual observations to navigational actions
\cite{chaplot2017gated, zhu2017target, zhu2017iccv, gupta_cvpr17, hermann2017grounded,
brahmbhatt_cvpr17,oh2017zero}.
We model our agents as similar pixel-to-action navigators. 
The key distinction in \eqa is how the goals are specified.
Visual navigation typically specifies agent goals either implicitly via the reward function
\cite{zhu2017iccv, gupta_cvpr17} (thus training a separate policy for each goal/reward),
or explicitly by conditioning on goal state representations~\cite{pfeiffer2017perception} including images of target objects \cite{zhu2017target}.
In contrast, \eqa specifies agent goals via language, which is inherently compositional
and renders training a separate policy for every task (question) infeasible. 

\xhdr{Situated Language Learning: Language + Action.}
%
Inspired by the classical work of Winograd~\cite{winograd_cogpsy72}, a number of
recent works have revisited grounded language learning by situating agents in
simple globally-perceived environments and tasking them with goals specified in natural language.
The form and structure of these goal specifications range from declarative programs
\cite{denil2017programmable}, to simple templated commands \cite{
yu2017deep, pmlr_v70_andreas17a}, to free-form natural language
instructions \cite{misra2017mapping, wang2016learning}.
One key distinction in \eqa, of course, is visual sensing -- the environment is only partially
observable, \ie the agent does not have access to the floor plan, object labels, attributes,
\etc, and must extract this information purely from first-person visual sensing.
%


\xhdr{Embodiment: Vision + Language + Action.}
Closest to \eqa are recent works that extend the situated language learning
paradigm to settings where agents' perceptions are local, purely visual, and change based
on their actions -- a setting we refer to as embodied language learning.

In concurrent and unpublished work, Hermann \etal~\cite{hermann2017grounded} and Chaplot \etal~%
\cite{chaplot2017gated} both develop embodied agents in simple game-like environments consisting
of 1-2 rooms and a handful of objects with variable color and shape. In both settings, agents were
able to learn to understand simple \myquote{go to $X$}/\myquote{pick up $X$} style commands where
$X$ would specify an object (and possibly some of its attributes). 
Similarly, Oh \etal \cite{oh2017zero} present embodied agents in a simple maze-world and task
them to complete a series of instructions. In contrast to these approaches, our \eqa environments
consist of multi-room homes (\myapprox8 per home) that are densely populated by a variety of objects
(\myapprox 54 unique objects per home). Furthermore, the instructions and commands in these works
are low-level and more closely relate to actions than the questions presented in \eqa.

\begin{figure*}[t]
\centering
\begin{subfigure}{0.25\textwidth}
\centering
\includegraphics[height=1.06in]{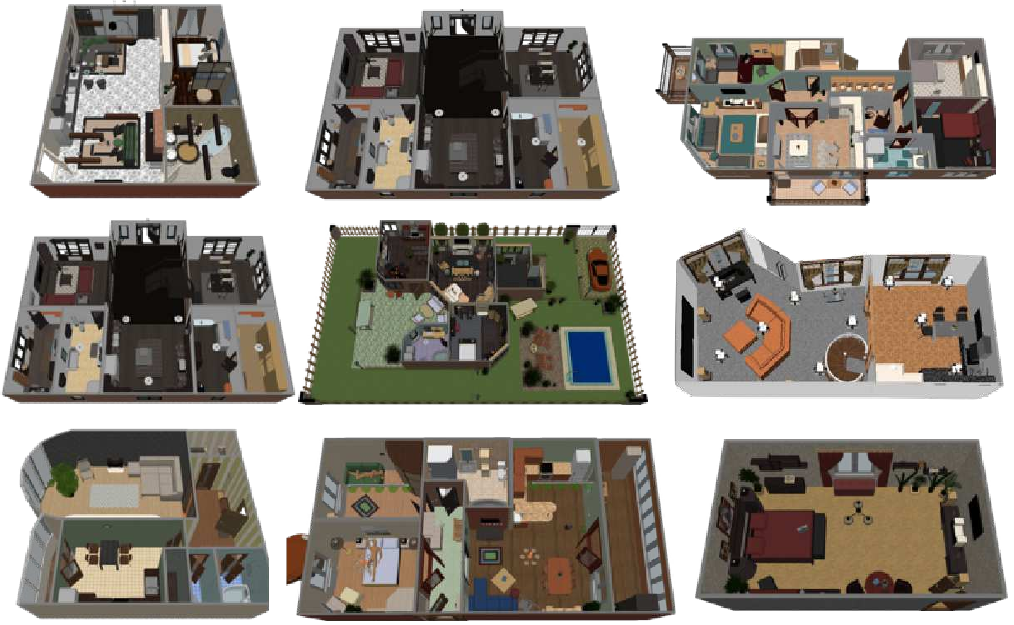}\\[2pt]
\caption{Sample Environments}
\label{fig:environ}
\end{subfigure}
\begin{subfigure}{0.20\textwidth}
\centering
\frame{\includegraphics[height=1.06in]{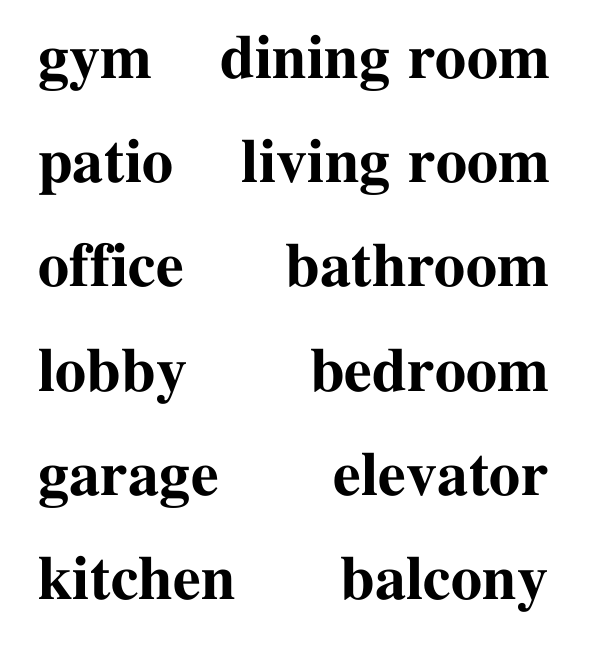}}\\[2pt]
\caption{Queryable Rooms}
\label{fig:rooms}
\end{subfigure}
\begin{subfigure}{0.54\textwidth}
\centering
\frame{\includegraphics[height=1.06in]{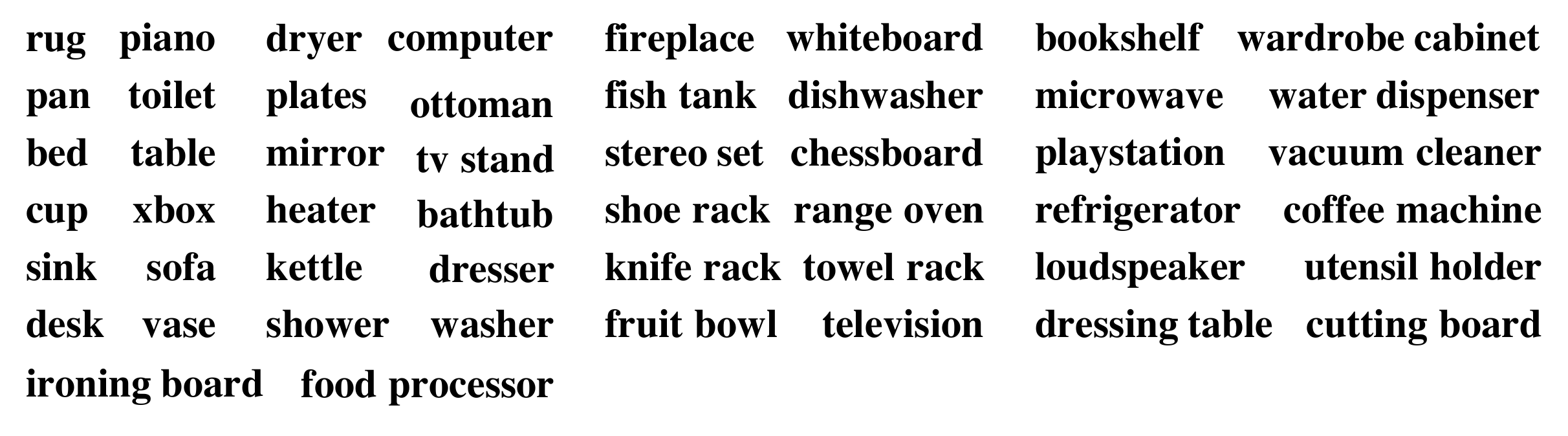}}\\[2pt]
\caption{Queryable Objects}
\label{fig:objects}
\end{subfigure}\\[2pt]
\caption{The \eqads dataset is built on a subset of the environments and objects from the SUNCG
\cite{song_cvpr17} dataset. We show (a) sample environments  and the (b) rooms and (c) objects
that are asked about in the \eqa task.} 
\end{figure*}

\xhdr{Interactive Environments.}
There are a number of interactive environments commonly used in the community, ranging from simple
2D grid-worlds (\eg XWORLD~\cite{yu2017deep}), to 3D game-like environments with limited realism
(\eg DeepMind Lab~\cite{Beattie16_deepmind_lab} or Doom~\cite{chaplot2017gated}), to more complex,
realistic environments (\eg AI2-THOR~\cite{zhu2017iccv} or Stanford 2D-3D-S~\cite{armeni_arxiv17}).
While realistic environments provide rich representations of the world, most
consist of only a handful of environments due to the high difficulty of their creation.
On the other hand, large sets of synthetic environments can be
programmatically generated; however, they typically lack realism (either in
appearance or arrangement).
In this work, we use the House3D~\cite{house3d} environment as it strikes a useful middle-ground
between simple synthetic and realistic environments. See \refsec{sec:envs} for more details.



\xhdr{Hierarchical Agents.}
We model our \eqa agents as deep hierarchical agents that decompose the overall control problem
such that a higher-level planner invokes lower-level controls to issue
primitive actions. Such hierarchical modeling has recently shown promise in the deep reinforcement
learning setting \cite{pmlr_v70_andreas17a,oh2017zero,TesslerGZMM16}. Our model also draws inspiration
from the work on Adaptive Computation Time models of Graves~\cite{graves_arxiv16}.
\section{\eqads Dataset: Questions In Environments}
\label{sec:qn-eng}

Having placed \eqa in context, we now dive deeper by outlining the
environments in which our agents are embodied and the questions
they must answer. We will publicly release the environments, our curated \eqads dataset,
and our code to aid research in this nascent area.
%

\subsection{House3D: Interactive 3D Environments}
\label{sec:envs}

We instantiate \eqa in House3D~\cite{house3d}, a recently introduced rich, interactive environment based on
3D indoor scenes from the SUNCG dataset~\cite{song_cvpr17}.
Concretely, SUNCG consists of synthetic 3D scenes with realistic room
and furniture layouts, manually designed using an online interior design interface (Planner5D \cite{planner5d}).
Scenes were also further `verified' as realistic by majority vote of three human annotators.
In total, SUNCG contains over $45$k environments with $49$k valid floors, $404$k rooms containing
$5$ million object instances of $2644$ unique objects from $80$ different categories.
House3D converts SUNCG from a static 3D dataset to a set of virtual environments,
where an agent (approximated as a cylinder 1 meter high)
may navigate under simple physical constraints (not being able to pass through walls or objects).
\figref{fig:environ} shows top-down views of sample environments.
Full details may be found in \cite{house3d}.


We build the \eqads dataset on a pruned subset of
environments from House3D.
First, we only consider environments for which all three SUNCG annotators consider the scene
layout realistic. Next, we filter out atypical environments such as those
lacking ground or those that are too small or large (only keeping houses with an internal area of
300-800$m^2$ covering at least $1/3$ the total ground area). Finally, we exclude
non-home environments by requiring at least one kitchen, living room, dining room, and bedroom.


\subsection{Question-Answer Generation}
\label{sec:questions}

We would like to pose questions to agents that test their abilities to ground
language, use common sense, reason visually, and navigate the environments.
For example, answering the question \myquote{What color is the car?} ostensibly requires
grounding the symbol \myquote{car}, reasoning that cars are typically outside,
navigating outside and exploring until the car is found, and visually inspecting its color.


We draw inspiration from
the CLEVR~\cite{johnson_cvpr17} dataset, and programmatically generate a dataset (\eqads) of grounded questions
and answers.
This gives us the ability to control the distribution of question-types and answers in the dataset, and
deter algorithms from exploiting dataset bias. 



\xhdr{Queryable Rooms and Objects.}
Figs.~\ref{fig:objects}, \ref{fig:rooms} show the queryable rooms (12) and objects (50) in \eqads.
We exclude objects and rooms from SUNCG that are obscure (\eg loggia rooms) or difficult
to resolve visually (\eg very small objects like light switches). We merge
some semantically similar object categories (\eg teapot, coffee kettle)
and singular vs plural forms of the same object type (\eg (books, book)) to reduce ambiguity.

\xhdr{Questions as Functional Programs.}
Each question in \eqads is represented as a functional program that can be executed on the environment
yielding an answer\footnote{or a response that the question is inapplicable (\eg referring to
objects not in the environment) or ambiguous (having multiple valid answers).}.
These functional programs are composed of a small set of elementary operations
($select(\cdot)$, $unique(\cdot)$, $query(\cdot)$, \etc)
that operate on sets of room or object annotations.

The number and the order of evaluation of these elementary operations defines a
\emph{question type} or template.
For instance, one question type in \eqads is the \texttt{location} template:
\begin{compactenum}[]
\item\texttt{location}: \myquote{What room is the <OBJ> located in?}
\end{compactenum}
where <OBJ> refers to one of the queryable objects.
The sequence of elementary operations for this question type is:
\vspace{-15pt}
\begin{align*}
select(objects) \rightarrow unique(objects) \rightarrow query(location).
\vspace{-5pt}
\end{align*}
The first function, $select(objects)$, gets all the object names from the environment. The second, $unique(objects)$, retains only the objects that have a single instance in the entire
house. The third, $query(location)$, generates a
question (by filling in the appropriate template) for each such object. The 2nd
operation, $unique(objects)$, is particularly important to generate unambiguous questions.
For instance, if there are two air conditioners in the house,
the question \myquote{What room is the air conditioner located in?}
is ambiguous, with potentially two different answers depending on which instance is being referred to.

\xhdr{Question Types.}
Associated with each question type is a template for generating a question about
the rooms and objects, their attributes and relationships.
We define nine question types and associated templates in \eqads: \\

{
\small
\begin{tabular}{@{}c@{} l@{\hspace{3pt}} p{2.1in}@{}}
\multirow{3}{*}{\rotatebox[origin=c]{90}{EQA v1} $\begin{dcases} \\ \\ \\ \\ \\  \end{dcases}$}
    &\texttt{location}: &\myquote{What room is the <OBJ> located in?} \\[2pt]
    &\texttt{color}: & \myquote{What color is the <OBJ>?} \\[2pt]
    &\texttt{color\_room}: & \myquote{What color is the <OBJ> in the <ROOM>?} \\[2pt]
    & \texttt{preposition}: & \myquote{What is <on/above/below/next-to> the <OBJ> in the <ROOM>?} \\[2pt]
    & \texttt{existence}: & \myquote{Is there a <OBJ> in the <ROOM>?} \\[2pt]
    & \texttt{logical}: & \myquote{Is there a(n) <OBJ1> and a(n) <OBJ2> in the <ROOM>?} \\[2pt]
    & \texttt{count}: & \myquote{How many <OBJs> in the <ROOM>?} \\[2pt]
    & \texttt{room\_count}: & \myquote{How many <ROOMs> in the house?} \\[2pt]
    & \texttt{distance}: & \myquote{Is the <OBJ1> closer to the <OBJ2> than to the <OBJ3> in the <ROOM>?}
\end{tabular}
}

The <ROOM> and <OBJ> tags above can be filled by any valid room or object
listed in \figref{fig:rooms} and \figref{fig:objects} respectively.  
Given these question templates, the possible answers are room names (\texttt{location}),
object names (\texttt{preposition}), yes/no (\texttt{existence}, \texttt{logical} and \texttt{distance}), color names
 (\texttt{color}) or numbers (\texttt{count}).

These questions test a range of agent abilities including object detection
(\texttt{existence}), scene recognition (\texttt{location}),
counting (\texttt{count}), spatial reasoning (\texttt{preposition}), color
recognition (\texttt{color}), and logical operators (\texttt{logic}).
Moreover, many of these questions require multiple capabilities: \eg,
answering a \texttt{distance} question requires recognizing the room
and objects as well as reasoning about their spatial relation.
Furthermore, the agent must do this by navigating the environment to find
the room, looking around the room to find the objects, and possibly
remembering their positions through time (if all three objects
are not simultaneously visible).

Different question types also require different degrees of navigation and memory. For instance,
\myquote{How many bedrooms in the house?} requires significant navigation (potentially exploring the
entire environment) and long-term memory (keeping track of the count), while a question
like \myquote{What color is the chair in the living room?} requires finding a single room, the living room, and
looking for a chair.

\eqads is easily extensible to include new elementary operations, question types, and
templates as needed to increase the difficulty of the task to match the development of
new models.
As a first step in this challenging space, our experiments focus on \eqads v1, which consists of 4 question types
-- \texttt{location}, \texttt{color}, \texttt{color\_room}, \texttt{preposition}.
One virtue of these questions is that there is a single target queried object (<OBJ>), which enables the
use of shortest paths from the agent's spawn location to the target as expert demonstrations for imitation learning
(details in \secref{sec:train}).
We stress that \eqads is not a static dataset, rather a curriculum of capabilities that we would
like to achieve in embodied communicating agents. 

\xhdr{Question-Answer Generation and Dataset Bias.}
In principle, we now have the ability to automatically generate all valid questions and their associated
answers for each environment by executing the functional programs on the environment's annotations
provided by SUNCG. However, careful consideration is needed to make sure the developed
dataset is balanced over question types and answers.



For each filled question template (\eg \myquote{What room is the refrigerator located in?}),
we execute its functional form on all associated environments in the dataset (\ie those
containing refrigerators) to compute the answer distribution for this question.
We exclude questions for which the normalized entropy of the answer distribution is below 0.5 --
\eg, an agent can simply memorize that refrigerators are almost always in kitchens,
so this question would be discarded.
We also exclude questions occurring in fewer than four environments as the
normalized entropy estimates are unreliable.

Finally, in order to benchmark performance of agents vs human performance on \eqads,
it is important for the questions to not be tedious or frustrating for humans to answer.
We do not ask \texttt{count} questions for objects with high
counts (>=5) or \texttt{distance} questions between object triplets without clear
differences in distance. We set these thresholds and room / object blacklists
manually based on our experience performing these tasks.

Complete discussion of the question templates, functional programs, elementary operations,
and various checks-and-balances can be found in the supplement.


\begin{figure}[h]
\vspace{3pt}
\begin{minipage}{0.57\columnwidth}
\centering
\resizebox{1.08\textwidth}{!}{
\setlength\tabcolsep{0pt}
\renewcommand{\arraystretch}{1.75}
\begin{tabular}{c c p{2.2cm} p{2cm}}
\toprule
  & Environments  & \centering {~~~~~Unique\newline\noindent~Questions} & {\centering ~~~~Total\newline\noindent~Questions}\\
\midrule
\texttt{train} &  643  & \centering 147  &   ~~~~4246 \\
\texttt{val} &  67  & \centering 104  &  ~~~~~506 \\
\texttt{test} &  57 & \centering 105 &   ~~~~~529 \\
\bottomrule
\end{tabular}}\\[10pt]
\end{minipage}%
\hspace{0.05\columnwidth}%
\begin{minipage}{0.36\columnwidth}
\centering
\includegraphics[width=0.97\textwidth,clip=true, trim=30px 10px 32px 5px]{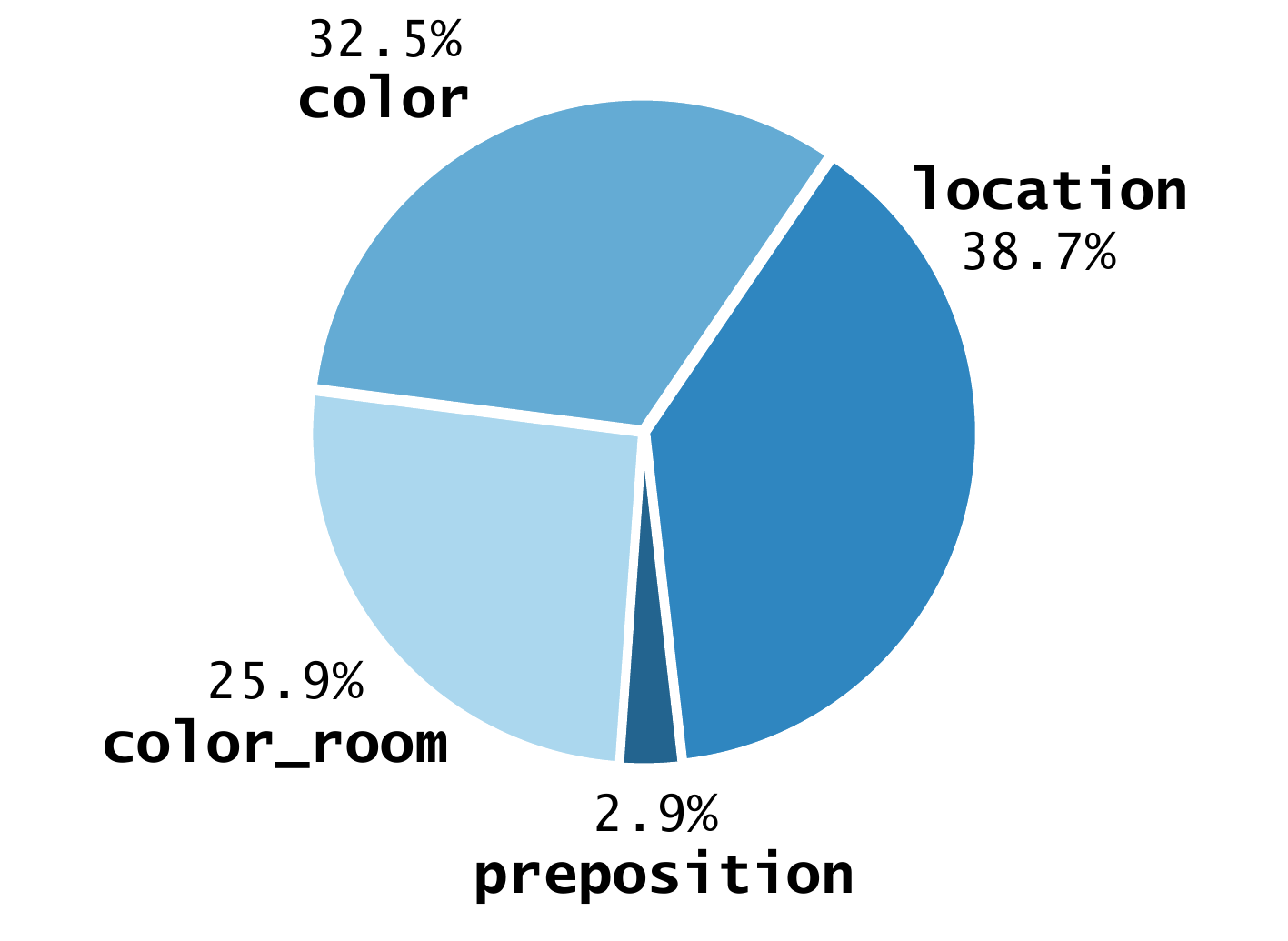}
\end{minipage}\\[3pt]
\captionof{figure}{Overview of the \eqads v1 dataset including dataset split statistics (left) and question type breakdown (right).}
\label{fig:eqastat}
\end{figure}

\begin{figure*}[t]
  \centering
\includegraphics[width=\textwidth]{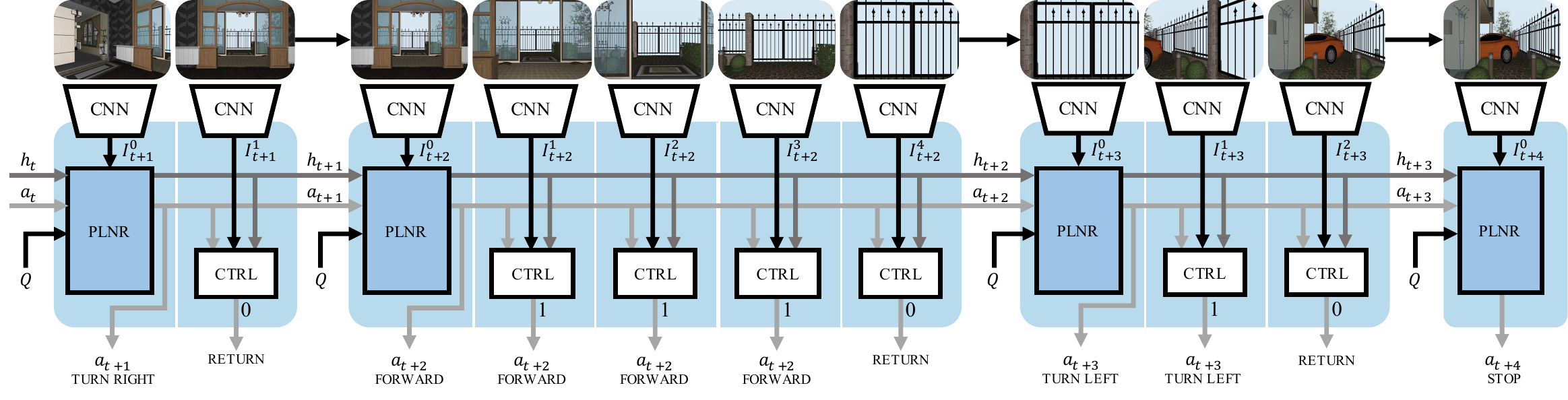}
\caption{Our Adaptive Computation Time (ACT) navigator splits the navigation task between a planner and a controller module. The planner
selects actions and the controller decides to continue performing that action for a variable number of time steps -- resulting in a decoupling
of direction (\myquote{turn left}) and velocity (\myquote{5 times}) and strengthening the long-term gradient flows of the planner module.
  }
\label{fig:approach}
\end{figure*}

\xhdr{\eqads v1 Statistics.}
The \eqads v1 dataset consists of over 5000 question across over 750 environments, referring to
a total of 45 unique objects in 7 unique room types.
The dataset is split into \texttt{train}, \texttt{val}, \texttt{test} such that there is
no overlap in environments across splits.
\reffig{fig:eqastat} shows the dataset splits 
and question type distribution.
Approximately 6 questions are asked per environment on average,
22 at most, and 1 at fewest. There are relatively few \texttt{preposition} questions
as many frequently occurring spatial relations are too easy to resolve without exploration and fail the
entropy thresholding. We will make \eqads v1 and the entire generation engine publicly available.

\vspace{\sectionReduceTop}
\section{A Hierarchical Model for \eqa}
\label{sec:approach}

We now introduce our proposed neural architecture for an \eqa agent.
Recall that the agent is spawned at a random location in the environment, receives a
question, and perceives only through a single egocentric RGB camera.
Importantly, unlike some prior work \cite{denil2017programmable,
yu2017deep, pmlr_v70_andreas17a,misra2017mapping, wang2016learning}, in \eqa, the agent does not
receive any global or structured representation of the environment (map, location, objects, rooms),
or of the task (the functional program that generated the question).

\textbf{Overview of the Agent.}
The agent has 4 natural modules -- vision, language, navigation, answering --
and is trained from raw sensory input (pixels and words)
to goal-driven multi-room indoor navigation to visual question answering.
The modules themselves are built up largely from conventional neural building blocks --
Convolutional Neural Networks (CNNs) and Recurrent Neural Networks (RNNs).
One key technical novelty in our model is the use of Adaptive Computation Time (ACT) RNNs
by Graves~\cite{graves_arxiv16}, which is an elegant approach for allowing RNNs to learn how many computational
steps to take between receiving an input and emitting an output by back-propagating through a `halting' layer.
We make use of this idea in our navigation module to cleanly separate the decision between --
direction (where to move, decided by a `planner') and velocity (how far to move, decided by a `controller').
\figref{fig:approach} illustrates the different modules in our agent, which we describe next.

\xhdr{Vision.}
Our agent takes egocentric $224{\times}224$ RGB images from the House3D renderer as input,
which we process with a CNN consisting of 4
$\{5{\times}5$ Conv, ReLU, BatchNorm, $2{\times}2$ MaxPool$\}$ blocks, producing
a fixed-size representation.

A strong visual system for \eqa should encode information about
object attributes (\ie colors and textures), semantics (\ie object
categories), and environmental geometry (\ie depth). As such, we
pretrain the CNN under a multi-task pixel-to-pixel prediction
framework -- treating the above CNN as an encoder network, we
train multiple network heads to decode the 1) original RGB values,
2) semantic class, and 3) depth of each pixel (which can be
obtained from the House3D renderer).
See the supplementary material for full model and training details.

\xhdr{Language.}
Our agents also receive questions which we encode with 2-layer LSTMs
with 128-dim hidden states.
Note that we learn \emph{separate} question encoders for the navigation and
answering modules --  as each may need to focus on different parts of the question.
For instance, in the question \myquote{What color is the chair in the kitchen?}, \myquote{color}
is irrelevant for navigation and \myquote{kitchen} matters little for question answering (once in the kitchen).

\xhdr{Navigation.}
We introduce a novel Adaptive Computation Time (ACT) navigator that decomposes navigation
into a `planner', that selects actions (forward, left, right),
and a `controller', that executes these primitive actions a variable number of times (1,2, \ldots) before returning control back to the planner.
%
%
%
Intuitively, this structure separates the intention of the agent
(\ie get to the other end of the room) from the series of
primitive actions required to achieve this directive (\ie \myquote{forward, forward, forward, ...}), and is reminiscent
of hierarchical RL approaches \cite{pmlr_v70_andreas17a,oh2017zero,TesslerGZMM16}.
This division also allows the planner to have variable time
steps between decisions, strengthening long-term gradient flows.  

Formally, let $t=1,2,\ldots,T$ denote planner timestamps, and $n = 0,1,2, \ldots N(t)$ denote the variable number of
controller steps. Let $I^n_t$ denote the encoding of the image observed at $t$-th planner-time and $n$-th controller-step.
We instantiate the planner as an LSTM. Thus, the planner maintains a hidden state $h^t$ (that is updated
only at planner timesteps), and samples an action $a_t \in \{$forward, turn-left, turn-right, stop-navigation$\}$:
\vspace{-2pt}
\begin{align}
a_t,~h_t \leftarrow \mathtt{PLNR}\left(h_{t-1},~I^0_t,~Q,~a_{t-1}\right),
\end{align}
where $Q$ is the question encoding.
After taking this action, the planner passes control to the controller, which considers the planner's state and the current frame to
decide to continue performing $a_t$ or to return control to the planner, \ie
\begin{align}
 \{0,1\} \ni c^n_t  \leftarrow \mathtt{CTRL}\left(h_t,a_t, I^{n}_t \right)
\end{align}
If $c^n_t = 1$ then the action $a_t$ repeats and $\mathtt{CTRL}$ is applied to the next frame.
Else if $c^n_t =0$ or a max of 5 controller-steps has been reached, control is returned to the planner.
We instantiate the controller as a feed-forward multi-layer perceptron with 1 hidden layer.
Intuitively, the planner encodes `intent' into the state encoding $h_t$ and the chosen action $a_t$,
and the controller keeps going until the visual input $I^n_t$ aligns with the intent of the planner.

\xhdr{Question Answering.}
After the agent decides to stop (or a max number of actions have been taken), the
question answering module is executed to provide an answer based on the sequence of frames $I^1_1, \ldots, I^n_T$
the agent has observed throughout its trajectory.
The answering module attends to each of the last five frame, computes an attention pooled visual
encoding based on image-question similarity, combines these with an LSTM encoding of the question,
and outputs a softmax over the space of $172$ possible answers. 



\subsection{Imitation Learning and Reward Shaping}
\label{sec:train}

We employ a two-stage training process. First, the navigation and answering
modules are independently trained using imitation/supervised learning on
automatically generated expert demonstrations of navigation. Second,
the entire architecture is jointly fine-tuned using policy gradients.

\xhdr{Independent Pretraining via Imitation Learning.}
Most questions that could be asked in \eqa do not have a natural `correct' navigation required to answer them.
As mentioned in \secref{sec:questions}, one virtue of \eqads v1 questions is that they contain
a single target queried object (<OBJ>). This allows us to use
the shortest path from the agent's spawn location to the target as an expert demonstration. 


The navigation module is trained to mimic the shortest path actions in a teacher forcing
setting - \ie, given the history encoding, question encoding, and the current frame, the
model is trained to predict the
action that would keep it on the shortest path. We use a
cross-entropy loss and train the model for 15 epochs. We find that even in this imitation
learning case, it is essential to train the navigator under a distance-based curriculum.
In the first epoch, we backtrack 10 steps from the target along the shortest path and initialize
the agent at this point with the full history of the trajectory from the spawned location. We step
back an additional 10 steps at each successive training epoch. We train for 15 epochs total with batch
size ranging from 5 to 20 questions (depending on path length due to memory limitations).

The question answering module is trained into predict the correct answer based on the question and the
frames seen on the shortest path. We apply standard cross-entropy training over 50 epochs
with a batch size of 20.

\xhdr{Target-aware Navigational Fine-tuning.}
While the navigation and answering modules that result from imitation learning perform
well on their independent tasks, they are poorly suited to dealing with each other. Specifically,
both modules are used to following the provided shortest path, but when in control the navigator
may generalize poorly and provide the question answerer with unhelpful views of target (if it finds
it at all). Rather than try to force the answering agent to provide correct answers from noisy
or absent views, we freeze it and fine-tune the navigator.

We provide two types of reward signals to the navigator: the question answering accuracy achieved
at the end of the navigation and a reward shaping~\cite{ng_icml99}
term that gives intermediate rewards for getting closer to the target.
Specifically, the answering reward is 5 if the agent chooses to stop and answers correctly and 0 otherwise.
The navigational reward for forward actions is 0.005 times the change in distance to
target object (there is no reward or penalty for turning).

We train the agent with REINFORCE \cite{williams1992simple} policy gradients with a running average
baseline for the answer reward. As in the imitation learning setting, we follow a curriculum of
increasing distance between spawn and target locations.

\textbf{Training details.}
All LSTMs are 2-layered with a $128$-d hidden state. We use Adam~\cite{kingma_iclr15} with
a learning rate of $10^{-3}$, and clamp gradients to $[-5,5]$. We incrementally load
environments in memory and use a batch size of $10$ both during the imitation
learning and REINFORCE fine-tuning stages.
One forward step corresponds to at most $0.25$ metres, and it takes 40 turns to turn $360^{\circ}$, \ie
one right or left turn action leads to $9^{\circ}$ change in viewing angle. Backward and strafe motions
are not allowed. We snap the continuous renderer space to a $1000{\times}1000$ grid to check for obstacles.
Our entire codebase will be publicly available.

\vspace{\sectionReduceTop}
\section{Experiments and Results}
\label{sec:results}
\vspace{\sectionReduceBot}

\begin{figure}[t]
\centering
\includegraphics[width=\columnwidth,clip=true, trim=3cm 8cm 12cm 2cm]{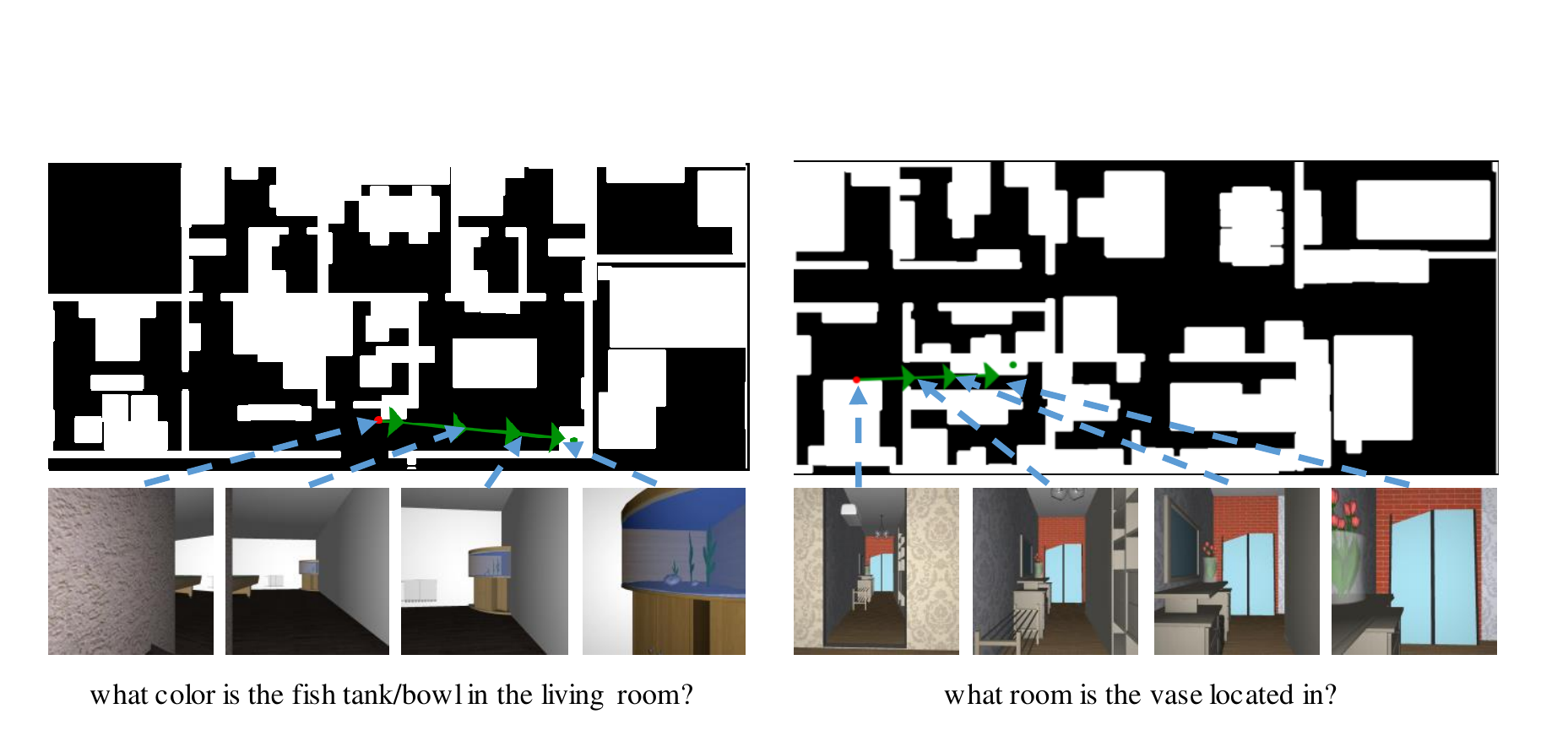}
\label{fig:top-down-viz}
\vspace{-10pt}
\caption{Sample trajectories from ACT+Q-RL agent projected on a floor plan (white areas are unoccupiable)
and on-path egocentric views.
The agent moves closer to already visible objects --
potentially improving its perception of the objects. Note that the floor plan is shown only for illustration and not available to the agents.}
\label{fig:top-down-viz}
\vspace{-15pt}
\end{figure}

\begin{table*}[t]
\setlength\tabcolsep{3pt}
\renewcommand{\arraystretch}{1.2}
\resizebox{\textwidth}{!}{
\begin{tabular}{c c l c c c l c c c l c c c l c c c l c c c l c c c l c c c}
& & \multicolumn{24}{c}{\textbf{Navigation}} &~~~ & \multicolumn{3}{c}{\textbf{QA}}\\
\cmidrule{4-26}\cmidrule{28-30}
&&& \multicolumn{3}{c}{$\mathbf{d_T}$} & & \multicolumn{3}{c}{$\mathbf{d_\Delta}$} & & \multicolumn{3}{c}{$\mathbf{d_{min}}$} & & \multicolumn{3}{c}{$\mathbf{\%r_T}$} & & \multicolumn{3}{c}{$\mathbf{\%r_{\dlsh}}$} & & \multicolumn{3}{c}{$\mathbf{\%stop}$} & & \multicolumn{3}{c}{$\mathbf{MR}$}\\
& && \scriptsize$T_{-10}$ & \scriptsize$T_{-30}$ & \scriptsize$T_{-50}$ &
& \scriptsize$T_{-10}$ & \scriptsize$T_{-30}$ & \scriptsize$T_{-50}$ &
 & \scriptsize$T_{-10}$ & \scriptsize$T_{-30}$ & \scriptsize$T_{-50}$ &
 & \scriptsize$T_{-10}$ & \scriptsize$T_{-30}$ & \scriptsize$T_{-50}$ &
 & \scriptsize$T_{-10}$ & \scriptsize$T_{-30}$ & \scriptsize$T_{-50}$ &
 & \scriptsize$T_{-10}$ & \scriptsize$T_{-30}$ & \scriptsize$T_{-50}$ &
 & \scriptsize$T_{-10}$ & \scriptsize$T_{-30}$ & \scriptsize$T_{-50}$ \\
 \toprule
\multirow{4}{*}{\rotatebox[origin=c]{90}{Baselines} $\begin{dcases} \\ \\ \\ \\  \end{dcases}$}
& Reactive && 2.09 & 2.72 & 3.14 && -1.44 & -1.09 & -0.31 && 0.29 & 1.01 & 1.82 && 50\% & 49\% & \textbf{47\%} && 52\% & 53\% & 48\% && - & - & - && 3.18 & 3.56 & 3.31\\
& LSTM && 1.75 & 2.37 & 2.90 && -1.10 & -0.74 & -0.07 && 0.34 & 1.06 & 2.05 && 55\% & 53\% & 44\% && 59\% & 57\% & 50\% && 80\% & 75\% & 80\% && 3.35 & 3.07 & 3.55 \\
& Reactive+Q && 1.58 & 2.27 & 2.89 && -0.94 & -0.63 & -0.06 && 0.31 & 1.09 & 1.96 && 52\% & 51\% & 45\% && 55\% & 57\% & \textbf{54\%} && - & - & - && 3.17 & 3.54 & 3.37 \\
& LSTM+Q && 1.13 & 2.23 & 2.89 && -0.48 & -0.59 & -0.06 && 0.28 & 0.97 & 1.91 && \textbf{63\%} & 53\% & 45\% && 64\% & 59\% & \textbf{54\%} && 80\% & 71\% & 68\% && 3.11 & 3.39 & 3.31 \\
\cmidrule{2-30}
\multirow{2}{*}{\rotatebox[origin=c]{90}{Us} $\begin{dcases} \\ \\ \end{dcases}$}
&ACT+Q && \textbf{0.46} & \textbf{1.50} & \textbf{2.74} && \textbf{0.16} & \textbf{0.15} & \textbf{0.12} && 0.42 & 1.42 & 2.63 && 58\% & 54\% & 45\% && 60\% & 56\% & 46\% && \textbf{100\%} & \textbf{100\%} & \textbf{100\%} && \textbf{3.09} & 3.13 & 3.25\\
&ACT+Q-RL && 1.67 & 2.19 & 2.86 && -1.05 & -0.52 & 0.01 && \textbf{0.24} & \textbf{0.93} & \textbf{1.94} && 57\% & \textbf{56\%} & 45\% && \textbf{65\%} & \textbf{62\%} & 52\% && 32\% & 32\% & 24\% && 3.13 & \textbf{2.99} & \textbf{3.22}\\
\cmidrule{2-30}
\multirow{2}{*}{\rotatebox[origin=c]{90}{Oracle} $\begin{dcases} \\ \\ \end{dcases}$}&HumanNav$^*$ && 0.81 & 0.81 & 0.81 && 0.44 & 1.62 & 2.85 && 0.33 & 0.33 & 0.33 && 86\% & 86\% & 86\% && 87\% & 89\% & 89\% && - & - & - && - & - & -\\
&ShortestPath+VQA && - & - & - && 0.85 & 2.78 & 4.86 && - & - & - && - & - & - && - & - & - && - & - & - && 3.21 & 3.21 & 3.21 \\[5pt]
\bottomrule
\end{tabular}}
\caption{Quantitative evaluation of \eqa agents on navigation and answering metrics for the \eqads v1 test set.
Ill-defined cells are marked with `-' because 1) reactive models don't have a stopping action,
2) humans pick a single answer from a drop-down list, so mean rank is not defined,
3) most distance metrics are trivially defined for shortest paths since they always end at the target object by design.}
\label{tab:results}
\end{table*}

The ultimate goal of an \eqa agent is to answer questions accurately. However,
it is important to disentangle success/failure at the intermediate task of navigation
from the ultimate downstream task of question answering.

\xhdr{Question Answering Accuracy.}
Our agent (and all baselines) produce a probability distribution over $172$ possible answers
(colors, rooms, objects). We report the mean rank ($\mathbf{MR}$) of the ground-truth answer in
the answer list sorted by the agent's beliefs, where the mean is computed over all \texttt{test} questions and environments.

\xhdr{Navigation Accuracy.}
We evaluate navigation performance on \eqads v1 by reporting the
distance to the target object at navigation termination $\left(\mathbf{d_T}\right)$,
change in distance to target from initial to final position $\left(\mathbf{{d_\Delta}}\right)$,
and the smallest distance to the target at any point in the episode $\left(\mathbf{{d_{min}}}\right)$.
\verify{All distances are measured in meters along the shortest path to the target.}
We also record the percentage of questions for which an agent either
terminates in $\left(\mathbf{\%r_T}\right)$
or ever enters $\left(\mathbf{\%r_{\dlsh}}\right)$ the room containing the target object(s).
Finally, we also report the percent of episodes in which agents choose to terminate navigation
and answer before reaching the maximum episode length $\left(\mathbf{\%stop}\right)$.
To sweep the difficulty of the task at test time, we spawn the agent 10, 30, or 50 actions
away from the target and report each metric for $T_{-10}$, $T_{-30}$, $T_{-50}$ settings.

\xhdr{Navigation Baselines.}
We compare our ACT navigator with a
number of sophisticated baselines and ablations.
\begin{compactenum}[\hspace{0pt}-]

	\item \textbf{Reactive CNN.} This is a feedforward network that uses the last-$n$
	frames to predict the next action. We tried $n= \{1, 3, 5, 10\}$ and report $n=5$, which worked best.
	Note that this is a target-agnostic baseline (\ie, is not aware of the question).
	The purpose of this baseline
	is to check whether simply memorizing frames from training environments generalizes
	to test (it does not).

	\item\textbf{Reactive CNN+Question.} This combines the frame representation (as above)
	with an LSTM encoding of the question to predict the next action. This is similar to the approach
	of \cite{zhu2017target}, with the difference that the goal is specified via a question encoding
	instead of a target image.


	Note that the action space for both the reactive baselines is $\{$forward, turn-left, turn-right$\}$,
	there is no stop token. At test time, the model is run for max no. of actions ($= 100$).

	\item\textbf{LSTM+Question.}
	The above two are memoryless navigators.
	This LSTM navigator takes as input the encodings of the question, current frame, and previous action,
	and predicts the next action. Note that these are identical inputs/outputs as our ACT navigator. The purpose
	of comparing to this ablation of our approach is to establish the benefit of our proposed planner-controller architecture.

	We also compare against an ablated version of this baseline without
	the question encoding as input (\textbf{LSTM}).

\end{compactenum}

\xhdr{Navigation Oracles.} We compare against two oracles:
\begin{compactenum}[\hspace{0pt}-]

	\item \textbf{HumanNav*} denotes goal-driven navigations by AMT workers remotely
	operating the agent
	($^*$ denotes that data human studies were conducted on a subset set of \texttt{test}).

	\item \textbf{ShortestPaths+VQA} denotes the question answering performance achieved
	by our answering module when fed in shortest path at \emph{test time}.
\end{compactenum}

\tableref{tab:results} shows the results of all baselines compared with our approach trained with just
imitation learning (ACT+Q) and our approach fine-tuned with RL (ACT+Q-RL)\footnote{\href{https://www.youtube.com/watch?v=gVj-TeIJfrk}{{\footnotesize\texttt{youtube.com/watch?v=gVj-TeIJfrk}}} shows example navigation and answer predictions by our agent.}.
We make a few key observations:

\begin{compactitem}

 	\item \textbf{All baselines are poor navigators.}
	All baselines methods have \emph{negative} $\mathbf{d_\Delta}$, \ie they end up \emph{farther}
	from the target than where they start. This confirms our intuition that \eqa is indeed a difficult problem. \\[-8pt]

	\item \textbf{Memory helps.}
	All models start equally far away from the target. Baselines augmented with memory
	(LSTM vs Reactive and LSTM-Q vs Reactive-Q)
	end closer to the target, \ie achieve smaller $\mathbf{d_T}$, than those without. \\[-8pt]

	\item \textbf{ACT Navigators performs best.}
	Our proposed navigator (ACT+Q) achieves the smallest distance to target at the end of navigation ($\mathbf{d_T}$),
	and the RL-finetuned navigator (ACT+Q-RL) achieves the highest answering accuracy.
	\\[-8pt]

	\item \textbf{RL agent overshoots.}
	Interestingly, we observe that while our RL-finetuned agent (ACT+Q-RL) gets closest to the target in its
	trajectory (\ie, achieves least $\mathbf{d_{min}}$) and enters the target room most often (\ie, achieves
	highest $\mathbf{\%r_{\dlsh}}$),
	it does \emph{not} end closest to the target (\ie,
	does not achieve highest $\mathbf{d_\Delta}$). These statistics and our qualitative analysis suggests
	that this is because RL-finetuned agents learn to explore, with a lower stopping rate ($\mathbf{\%stop}$),
	and often overshoot the target. This is
	consistent with observations in literature~\cite{dipendra_emnlp17}. In \eqa, this overshooting behavior
	does not hurt the question answering accuracy because the answering module can attend to frames
	along the trajectory. 
	This behavior can be corrected by including a small 	negative reward for each action.
	\\[-8pt]

	\item \textbf{Shortest paths are not optimal for VQA.}
	A number of methods outperform ShortestPath+VQA in terms of answering accuracy. This is because
	while the shortest path clearly takes an agent to the target object, it may not provide the best vantage
	to answer the question. In future work, these may be improved by ray tracing methods to 
	appropriately frame of the target object at termination.

\end{compactitem}

\section{Conclusion}
\label{sec:conc}

We present Embodied Question Answering (\eqa) -- a new AI task where an agent is spawned at a random location in a 3D environment and asked a question. In order to answer, the agent
must first intelligently navigate to explore the environment, gather information through first-person (egocentric) vision, and then answer the question. We develop a novel neural hierarchical model that decomposes navigation into a `planner' -- that selects actions or a direction -- and a `controller' -- that selects a velocity and executes the primitive actions a variable number of times -- before returning control to the planner. We initialize the agent via imitation learning and fine-tune it using reinforcement learning for the goal of answering questions. We develop evaluation protocols for \eqa, and evaluate our agent in the House3D virtual environment. Additionally, we collect human demonstrations by connecting workers on Amazon Mechanical Turk to this environment to remotely control an embodied agents. All our code, data, and infrastructure will be made publicly available.

\section{Acknowledgements}
\label{sec:ack}

We are grateful to the developers of PyTorch~\cite{pytorch} for building an excellent framework.
We thank Yuxin Wu for help with the House3D environment.
This work was funded in part by NSF CAREER awards to DB and DP, ONR YIP awards
to DP and DB, ONR Grant N00014-14-1-0679 to DB, ONR Grant N00014-16-1-2713 to DP,
an Allen Distinguished Investigator award to DP from the Paul G. Allen Family Foundation,
Google Faculty Research Awards to DP and DB, Amazon Academic Research Awards to DP and DB,
AWS in Education Research grant to DB, and NVIDIA GPU donations to DB. The views and
conclusions contained herein are those of the authors and should not be interpreted as
necessarily representing the official policies or endorsements, either expressed
or implied, of the U.S. Government, or any sponsor.


\clearpage

\appendix
\section*{Appendix Overview}




This supplementary document is organized as follows:

\begin{itemize}

\item \refsec{sec:qengine} presents the question-answer generation engine in detail,
including functional programs associated with questions, and checks and balances in
place to avoid ambiguities, biases, and redundancies.

\item \refsec{sec:cnn} describes the CNN models that serve as the vision module
for our \eqa model. We describe the model architecture, along with the
training details, quantitative as well as qualitative results.

\item \refsec{sec:answerer} describes the answering module in our agent.

\item \refsec{sec:humannav} reports machine question answering performance conditioned on human navigation paths
(collected via human studies on Amazon Mechanical Turk).

\item Finally, \href{https://www.youtube.com/watch?v=gVj-TeIJfrk}{{\footnotesize\texttt{youtube.com/watch?v=gVj-TeIJfrk}}}
shows example navigation and answer predictions by our agent.
\end{itemize}


\begin{table*}
\centering
\renewcommand*{\arraystretch}{1.25}
\resizebox{\textwidth}{!}{
\begin{tabular}{l l}

\hline
\textbf{Template} & \textbf{Functional Form} \\
\hline

\small{\texttt{location}} & \small{$select(objects) \rightarrow unique(objects) \rightarrow blacklist(location) \rightarrow query(location)$} \\

\small{\texttt{color}} & \small{$select(objects) \rightarrow unique(objects) \rightarrow blacklist(color)
	\rightarrow query(color)$} \\

\small{\texttt{color\_room}} & \small{$select(room) \rightarrow unique(room) \rightarrow select(objects)
	\rightarrow unique(objects) \rightarrow blacklist(color) \rightarrow query(color\_room)$} \\

\small{\texttt{preposition}} & \small{$select(room) \rightarrow unique(room) \rightarrow select(objects)
	\rightarrow unique(objects) \rightarrow blacklist(preposition) \rightarrow relate()
	\rightarrow query(preposition)$} \\

\small{\texttt{existence}} & \small{$select(room) \rightarrow unique(room) \rightarrow select(objects) \rightarrow blacklist(existence) \rightarrow query(exist)$} \\

\small{\texttt{logical}} & \small{$select(room) \rightarrow unique(room) \rightarrow select(objects) \rightarrow blacklist(existence) \rightarrow query(logical)$} \\

\small{\texttt{count}} & \small{$select(room) \rightarrow unique(room) \rightarrow select(objects)
	\rightarrow blacklist(count) \rightarrow query(count)$} \\

\small{\texttt{room\_count}} & \small{$select(room) \rightarrow query(room\_count)$} \\

\small{\texttt{distance}} & \small{$select(room) \rightarrow unique(room) \rightarrow select(objects) \rightarrow unique(objects) \rightarrow blacklist(distance) \rightarrow distance() \rightarrow query(distance)$} \\
\hline

\end{tabular}}\\[0pt]
\caption{Functional forms of all question types in the \eqads dataset}
\label{tab:func-forms}
\vspace{-10pt}
\end{table*}

\section{Question-Answer Generation Engine}
\label{sec:qengine}

Recall that each question in \eqads is represented as a functional program that can be executed on the environment
yielding an answer\footnote{or a response that the question is inapplicable (\eg referring to
objects not in the environment) or ambiguous (having multiple valid answers).}.
In this section, we describe this process in detail.
In the descriptions that follow, an `entity' can refer to either a queryable room or
a queryable object from the House3D~\cite{house3d} environment.

\xhdr{Functional Forms of Questions.}
The functional programs are composed of elementary operations described below:

\begin{enumerate}
	\item $select(entity)$: Fetches a list of entities from the environment. This operation
	is similar to the `select' query in relational databases.

	\item $unique(entity)$: Performs filtering to retain entities that are unique (\ie occur exactly once).
	For example, calling 	$unique(rooms)$ on the set of rooms $[ \myquote{living\_room}, \myquote{bedroom}, \myquote{bedroom}]$
	for a given house will return $[ \myquote{living\_room} ]$.

	\item $blacklist(template)$: This function operates on a list of object entities and
	filters out a pre-defined list of objects that are blacklisted for the
	given template. We do not ask questions of a given template type
	corresponding to any of the blacklisted objects. For example, if the
	blacklist contains the objects $ \{\myquote{column}, \myquote{range\_hood},
	\myquote{toy}\} $ and the objects list that the function receives
	is $\{\myquote{piano}, \myquote{bed}, \myquote{column}\}$, then
	the output of the $blacklist()$ function is:
	$\{\myquote{piano}, \myquote{bed}\}$

	\item $query(template)$: This is used to generate the questions strings
	for the given template on the basis of the entities that it receives.
	For example, if $query(location)$ receives the following set of object
	entities as input: $[\myquote{piano}, \myquote{bed}, \myquote{television}]$, it produces $3$
	question strings of the form: \textit{what room is the <OBJ> located in?}
	where \textit{<OBJ>} $ = \{\myquote{piano}, \myquote{bed}, \myquote{television}\}$.

	\item $relate()$: This elementary operation is used in the functional form for
	preposition questions. Given a list of object entities, it returns a subset of
	the pairs of objects that have a $\{\myquote{on}, \myquote{above}, \myquote{under}, \myquote{below},
	\myquote{next to}\}$ spatial relationship between them.

	\item $distance()$: This elementary operation is used in the functional form for
	distance comparison questions. Given a list of object entities, it returns triplets
	of objects such that the first object is closer/farther to the anchor object than
	the second object.

\end{enumerate}

Having described the elementary operations that make up our functional forms, the
explanations of the functional forms for each question template is given below.
We categorize the question types into $3$ categories based on the objects and the rooms
that they refer to.

\begin{enumerate}
	\item \textbf{Unambiguous Object}: There are certain question types that inquire about an
	object that must be unique and unambiguous throughout the environment. Examples of such
	question types are \texttt{location} and \texttt{color}. For example, we should ask \myquote{what room is the piano
	located in?} if there is  only a single instance of a \myquote{piano} in the environment. Hence, the
	functional forms for location and color have the following structure:\\

	\begin{minipage}{\columnwidth}
	\centering
	$select(objects) \rightarrow unique(objects) \rightarrow query(location/color)$.
	\end{minipage}\\

	$select(objects)$ gets the list of all objects in the house and the $unique(objects)$ only retains objects
	that are unique, thereby ensuring unambiguity. The $query(template)$ function prepares the question
	string by filling in the slots in the template string.

	\item \textbf{Unambiguous Room + Unambiguous Object}: In continuation of the above, there is
	another set of question types that talk about objects in rooms where in addition to the objects
	being unique, the rooms should also be unambiguous. Examples of such question types include \texttt{color\_room},
	\texttt{preposition}, and \texttt{distance}. The additional unambiguity constraint on the room is because the question
	\myquote{what is next to the bathtub in the bathroom?} would become ambiguous if there are two or more bathrooms
	in the house. The functional forms for such types are given by the following structure:\\

	\begin{minipage}{\columnwidth}
	\centering
	$select(rooms) \rightarrow unique(rooms) \rightarrow select(objects) \rightarrow unique(objects)
	\rightarrow query(template)$.
	\end{minipage}\\

	The first two operations in the sequence result in a list of unambiguous rooms
	whereas the next two result in a list of unambiguous objects in those rooms. Note that when
	$select(\cdot)$ appears as the first operation in the sequence (\ie, select operates on an empty list),
	it is used to fetch the set of rooms or objects across the entire house. However, in this case,
	$select(object)$ operates on a set of rooms (the output of $select(rooms) \rightarrow unique(rooms) \rightarrow$),
	so it returns the set of objects found in those specific rooms (as opposed to fetching objects
	across all rooms in the house).

	\item \textbf{Unambiguous Room}: The final set of question types are the ones where the rooms need
	to be unambiguous, but the objects in those rooms that are being referred to do not. Examples of
	such question types are: \texttt{existence}, \texttt{logical}, and \texttt{count}. It is evident that for asking about the
	existence of objects or while counting them, we do not require the object to have only a single
	instance. \myquote{Is there a television in the living room?} is a perfectly valid question, even if there
	are multiple televisions in the living room (provided that there is a single living room in the house).
	The template structure for this is a simplified version of (2):\\

	\begin{minipage}{\columnwidth}
	\centering
	$select(rooms) \rightarrow unique(rooms) \rightarrow select(objects) \rightarrow query(template)$.
	\end{minipage}\\

	Note that we have dropped $unique(objects)$ from the sequence as we no longer require that condition to hold true.

\end{enumerate}
See Table \ref{tab:func-forms} for a complete list of question functional forms.

\begin{table*}[t]
    \centering
    \begin{minipage}{.55\linewidth}
      \centering
        \begin{tabular}{cccc}
        	\hline
             & Pixel Accuracy & Mean Pixel Accuracy & Mean IOU \\
             \hline
             single & 0.780 & 0.246 & 0.163 \\
             hybrid & 0.755 & 0.254 & 0.166 \\
             \hline
        \end{tabular}
        \caption*{(a)	 Segmentation}
    \end{minipage}\hfill%
    \begin{minipage}{.2\linewidth}
      \centering
        \begin{tabular}{lc}
            \hline
            & Smooth-$\ell_1$ \\
            \hline
            single & 0.003 \\
            hybrid & 0.005 \\
            \hline
        \end{tabular}
        \caption*{(b) Depth}
    \end{minipage}\hfill%
    \begin{minipage}{.2\linewidth}
      \centering
        \begin{tabular}{lc}
            \hline
            & Smooth-$\ell_1$ \\
            \hline
            single & 0.003 \\
            hybrid & 0.003 \\
            \hline
        \end{tabular}
        \caption*{(c) Autoencoder}
    \end{minipage}\\[0pt]

    \caption{Quantitative results for the autoencoder, depth estimation, and semantic segmentation
    heads of our multi-task perception network. All metrics are reported on a held out validation set.}
    \label{tab:cnn}
    \vspace{-10pt}
\end{table*}

\xhdr{Checks and balances.}
Since one of our goals is to benchmark performance of our agents with human performance, we want
to avoid questions that are cumbersome for a human to navigate for or to answer. Additionally,
we would also like to have a balanced distribution of answers so that the agent is not able to
simply exploit dataset biases and answer questions effectively without exploration.
This section describes in detail the various checks that we have in place to ensure
these properties.

\begin{enumerate}
	\item \textbf{Entropy+Frequency-based Filtering}:
	It is important to ensure that the distribution of answers for a question is not too `peaky',
	otherwise the mode guess from this distribution will do unreasonably well as a baseline.
	Thus, we compute the normalized entropy of the distribution
	of answers for a question. We drop questions where the normalized
	entropy is below $0.5$. Further, we also drop questions that occur in less
	than $4$ environments because the entropy counts for low-frequency
	questions are not reliable.

	\item \textbf{Object Instance Count Threshold}: We do not ask counting questions (\texttt{room\_count} and \texttt{count})
	when the answer is greater than $5$, as they are tedious for humans.

	\item \textbf{Object Distance Threshold}: We consider triplets of objects within
    a room consisting of an anchor object, such that the difference of distances
    between two object-anchor pairs is at least $2$ metres.
    This is to avoid ambiguity in \myquote{closer}/\myquote{farther} object distance comparison questions.

	\item \textbf{Collapsing Object Labels}: Object types that are visually very similar
	(\eg \myquote{teapot} and \myquote{coffee\_kettle}) or semantically hierarchical in relation (\eg \myquote{bread} and \myquote{food}) introduce
	unwanted ambiguity. In these cases we collapse the object labels to manually selected labels
	(\eg (\myquote{teapot}, \myquote{coffee\_kettle}) $\rightarrow$ \myquote{kettle} and (\myquote{bread}, \myquote{food}) $\rightarrow$ \myquote{food}).

	\item \textbf{Blacklists}:\\[-17pt]
	\begin{itemize}
		\item \textbf{Rooms}: Some question types in the \eqads dataset have room names
		in the question string (e.g. \texttt{color\_room}, \texttt{exist}, \texttt{logical}). We do not generate such questions
		for rooms that have obscure or esoteric names such as \myquote{loggia}, \myquote{freight elevator}, \myquote{aeration} \etc or
		names from which the room being referred might not be immediately obvious \eg \myquote{boiler room}, \myquote{entryway} \etc.

		\item \textbf{Objects}: For each question template, we maintain a list of objects that are
		not to be included in questions. These are either tiny objects or whose name descriptions are too
		vague e.g. \myquote{switch} (too small), \myquote{decoration} (not descriptive enough), \myquote{glass} (transparent),
		\myquote{household appliance} (too vague). These blacklists are manually defined based on our experiences performing these tasks.
	\end{itemize}
\end{enumerate}

\newcommand{\subf}[1]{%
  {\small\begin{tabular}[t]{@{}c@{}}
  #1
  \end{tabular}}%
}

\section{CNN Training Details}
\label{sec:cnn}

\begin{figure*}
\centering
    \def\evlen{1.0\textwidth}
    \def\evr{0.1\textwidth}
\setlength{\tabcolsep}{2pt}
\resizebox{0.85\textwidth}{!}{
\begin{tabular}{cc@{\hskip 1.5em}cc@{\hskip 1.5em}cc}
\hline \\[-0.2cm]
 \textbf{GT RGB} & \textbf{Pred. RGB} & \textbf{GT Depth} & \textbf{Pred.~Depth} & \textbf{GT Seg.} & \textbf{Pred.~Seg.} \\[0.05cm]
\hline \\[-0.1cm]

\subf{\includegraphics[width=\evr]{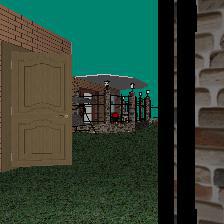}}
&
\subf{\includegraphics[width=\evr]{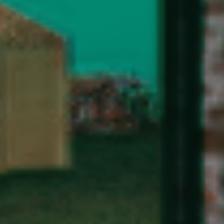}}
&
\subf{\includegraphics[width=\evr]{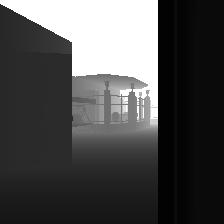}}
&
\subf{\includegraphics[width=\evr]{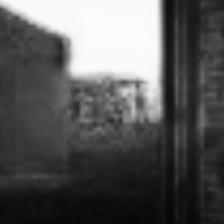}}
&
\subf{\includegraphics[width=\evr]{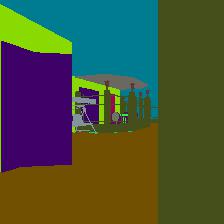}}
&
\subf{\includegraphics[width=\evr]{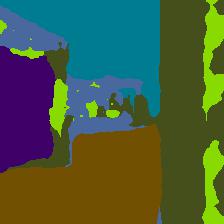}}
\\

\subf{\includegraphics[width=\evr]{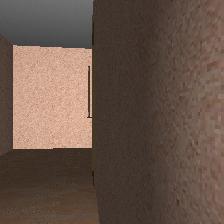}}
&
\subf{\includegraphics[width=\evr]{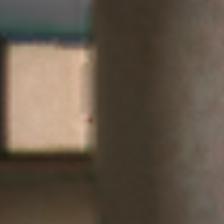}}
&
\subf{\includegraphics[width=\evr]{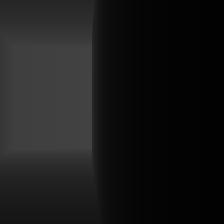}}
&
\subf{\includegraphics[width=\evr]{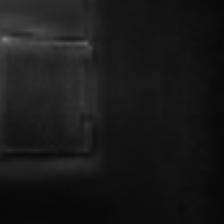}}
&
\subf{\includegraphics[width=\evr]{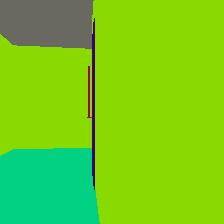}}
&
\subf{\includegraphics[width=\evr]{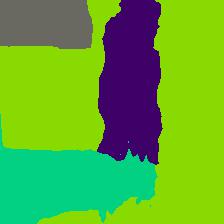}}
\\

\subf{\includegraphics[width=\evr]{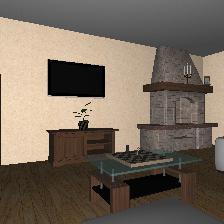}}
&
\subf{\includegraphics[width=\evr]{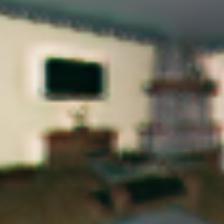}}
&
\subf{\includegraphics[width=\evr]{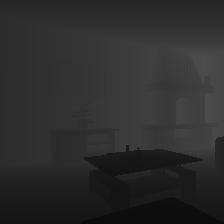}}
&
\subf{\includegraphics[width=\evr]{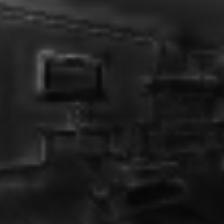}}
&
\subf{\includegraphics[width=\evr]{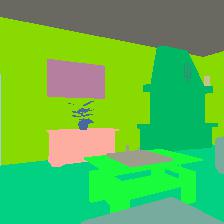}}
&
\subf{\includegraphics[width=\evr]{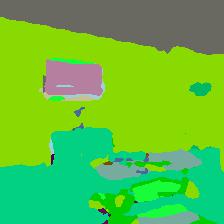}}
\\

\subf{\includegraphics[width=\evr]{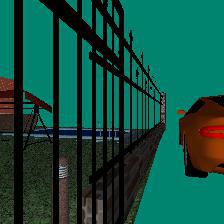}}
&
\subf{\includegraphics[width=\evr]{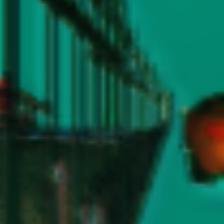}}
&
\subf{\includegraphics[width=\evr]{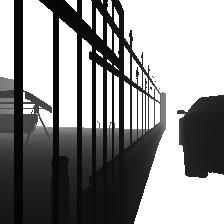}}
&
\subf{\includegraphics[width=\evr]{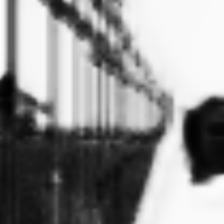}}
&
\subf{\includegraphics[width=\evr]{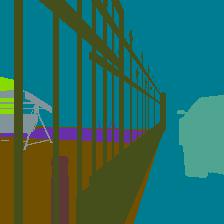}}
&
\subf{\includegraphics[width=\evr]{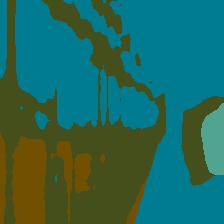}}
\\

\end{tabular}}\\
\caption{Some qualitative results from the hybrid CNN. Each row represents
an input image. For every input RGB image, we show the reconstruction from the
autoencoder head, ground truth depth, predicted depth as well as ground truth segmentation and predicted segmentation maps.}
\label{fig:cnn-qual}
\vspace{-10pt}
\end{figure*}

The CNN comprising the visual system for our \eqa agents is trained under a
multi-task pixel-to-pixel prediction framework. We have an encoder network
that transforms the egocentric RGB image from the House3D renderer~\cite{house3d} to a
fixed-size representation. We have $3$ decoding heads
that predict 1) original RGB values (\ie an autoencoder), 2) semantic class, and 3)
depth for each pixel. The information regarding semantic class and depth of
every pixel is available from the renderer. The range of depth values for every
pixel lies in the range $[0, 1]$ and the segmentation is done over 191 classes.

\xhdr{Architecture.}
The encoder network has 4 Conv blocks, comprising of $5 \times 5$ Conv filters,
ReLU, BatchNorm and $2 \times 2$ MaxPool. Each of the $3$ decoder branches
of our network upsample the
encoder output to the spatial size of the original input image. The number
of channels in the output of the decoder depends on the task head -- 191, 1
and 3 for the semantic segmentation, depth and autoencoder branches respectively.
The upsampling is done using bilinear interpolation.
The architecture also has skip connections from the 2nd and the 3rd Conv layers.

\xhdr{Training Details.}
We use cross-entropy loss to train the segmentation branch of our hybrid network.
The depth and autoencoder branches are trained using the Smooth-$\ell_1$ loss.
The total loss is a linear combination of the 3 losses, given by
$overall\_loss = seg\_loss + 10 {\times} depth\_loss + 10 {\times} reconstruction\_loss$.
We use the Adam optimizer with a learning rate of $10^{-3}$ and a batch size of 20.
The hybrid network is trained for a total of 5 epochs on a dataset of $100k$
RGB training images from the renderer.

\xhdr{Quantitative Results.}
Table \ref{tab:cnn} shows some quantitative results. For each of the $3$
different decoding heads of our multi-task CNN, we report results on the
validation set for two settings - when the network is trained for all tasks at once
(hybrid) or each task independently (single).
For segmentation, we report the overall pixel accuracy, mean pixel accuracy
(averaged over all $191$ classes) and the mean IOU (intersection over union).
For depth and autoencoder, we report the Smooth-$\ell_1$ on the validation set.

\xhdr{Qualitative Results.}
Some qualitative results on images from the validation set for
segmentation, depth prediction and autoencoder reconstruction
are shown in Figure \ref{fig:cnn-qual}.

\section{Question Answering Module}
\label{sec:answerer}

\begin{figure}
\centering
\includegraphics[width=\columnwidth]{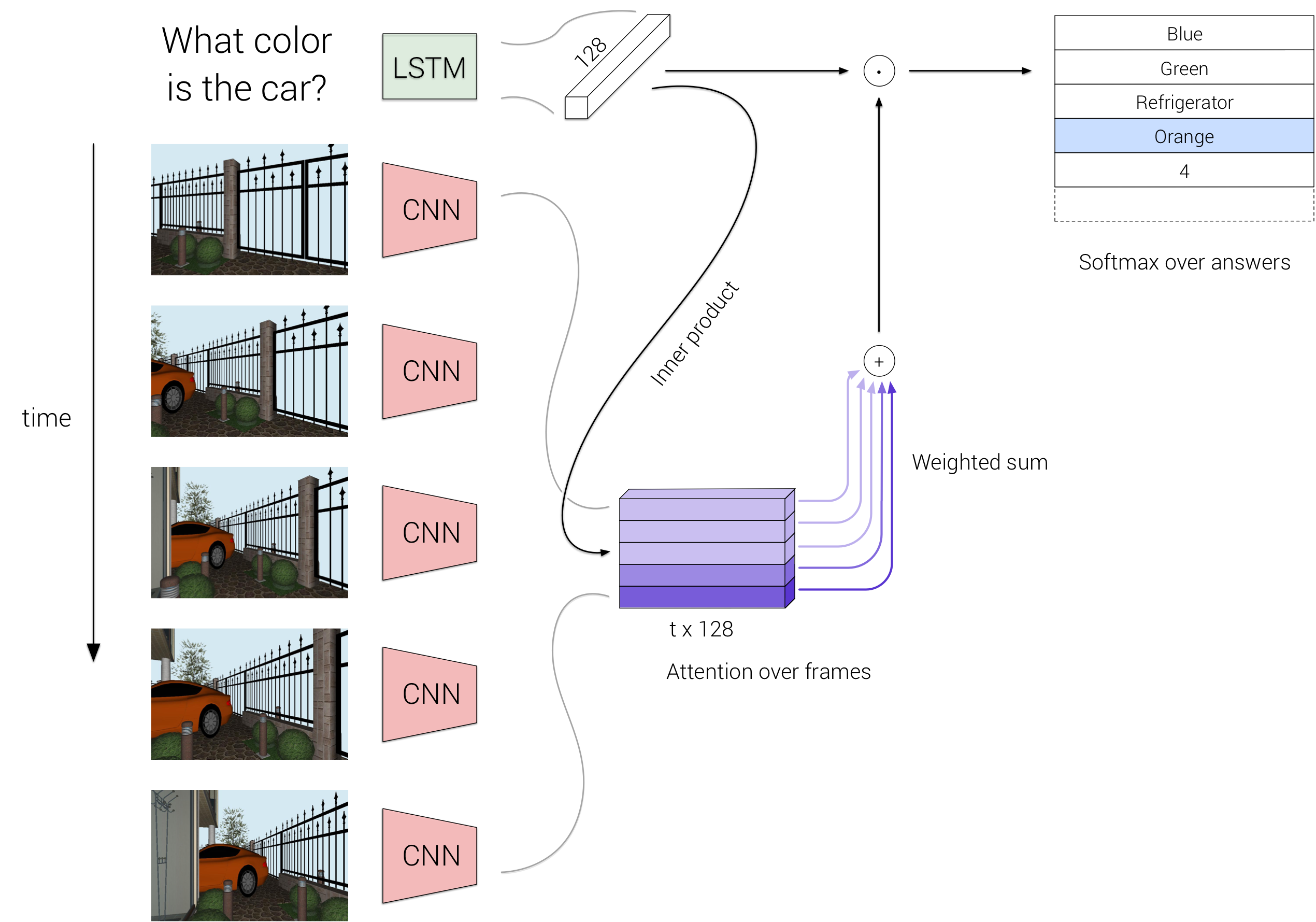}
\vspace{5pt}
\captionof{figure}{Conditioned on the navigation frames and question,
the question answering module computes dot product attention over the last five frames,
and combines attention-weighted combination of image features with question encoding
to predict the answer.
}
\label{fig:answerer}
\vspace{-10pt}
\end{figure}

The question answering module predicts the agents' beliefs over the answer given the agent's navigation. 
It first encodes the question with an LSTM, last five frames of the navigation each with a CNN, and 
then computes dot product attention over the five frames to pick 
the most relevant ones. Next, it combines attention-weighted sum of image features 
with the question encoding to predict a softmax distribution over answers. See \figref{fig:answerer}.

\vspace{10pt}
\section{Human Navigation $+$ Machine QA}
\label{sec:humannav}

\begin{figure*}
\centering
\includegraphics[width=0.8\textwidth]{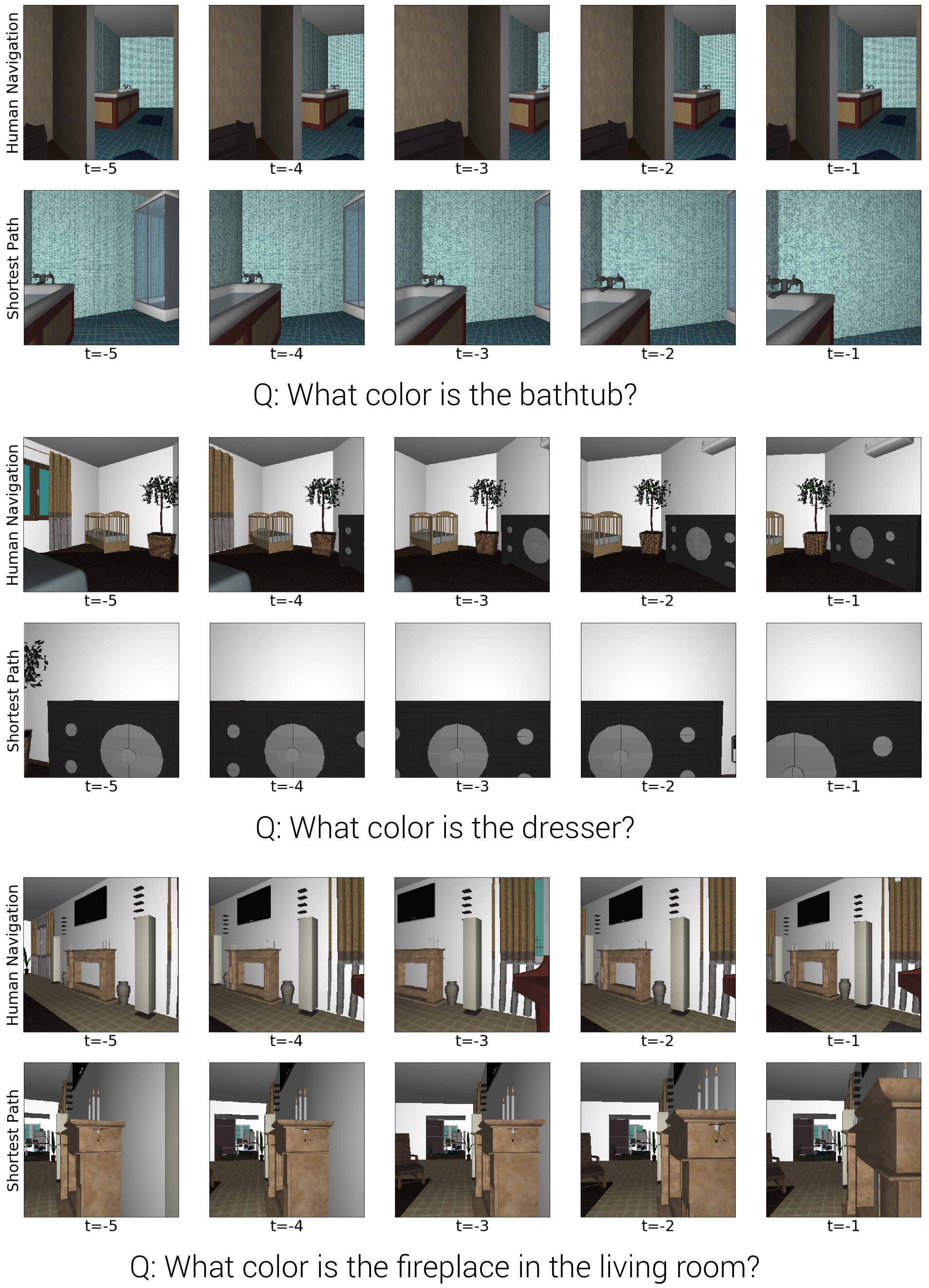}
\captionof{figure}{Examples of last five frames from human navigation vs. shortest path.}
\label{fig:human-shortnav}
\end{figure*}

In order to contrast human and shortest-path navigations with respect to question answering,
we evaluate our QA model on the last $5$ frames of human navigations collected through our
Amazon Mechanical Turk interface. We find the mean rank of the ground truth answer to be 3.51
for this setting (compared to 3.26 when computed from shortest-paths). We attribute this difference 
primarily to a mismatch between the QA system training on shortest paths and testing on 
human navigations. While the shortest paths typically end with the object of interest filling the
majority of the view, humans tend to stop earlier as soon as the correct answer can be discerned.
As such, human views tend to be more cluttered and pose a more difficult task for the QA module. 
\figref{fig:human-shortnav} highlights this difference by contrasting the last $5$ frames from 
human and shortest-path navigations across three questions and environments.

\begin{figure*}
\centering
	\def\evlen{1.0\textwidth}
	\def\evr{0.16\textwidth}
\setlength{\tabcolsep}{0pt}
\resizebox{\textwidth}{!}{    
\begin{tabular}{c c c c c }

\rotatebox{90}{\hspace{5pt}\textbf{air conditioner}}
&
\subf{\includegraphics[width=\evr]{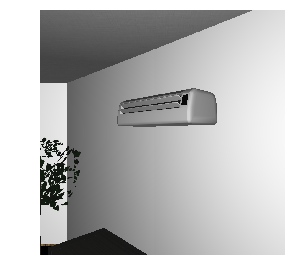}}
&
\subf{\includegraphics[width=\evr]{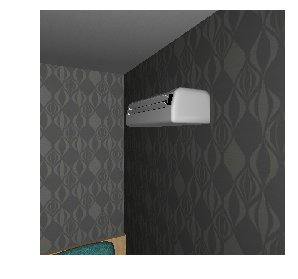}}
&
\subf{\includegraphics[width=\evr]{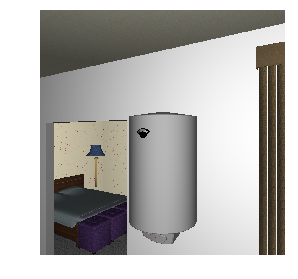}}
&
\subf{\includegraphics[width=\evr]{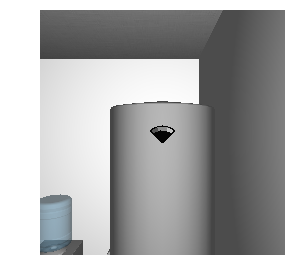}}
\\

\rotatebox{90}{\hspace{25pt}\textbf{candle}}
&
\subf{\includegraphics[width=\evr]{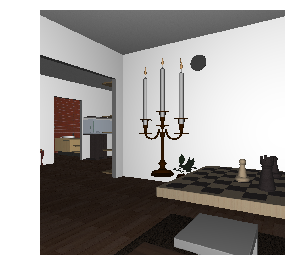}}
&
\subf{\includegraphics[width=\evr]{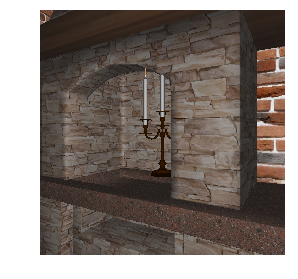}}
&
\subf{\includegraphics[width=\evr]{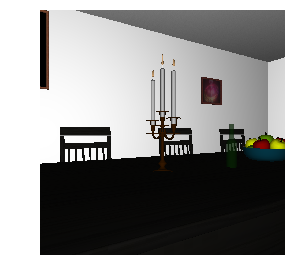}}
&
\subf{\includegraphics[width=\evr]{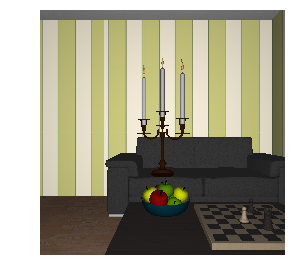}}
\\

\rotatebox{90}{\hspace{15pt}\textbf{pedestal fan}}
&
\subf{\includegraphics[width=\evr]{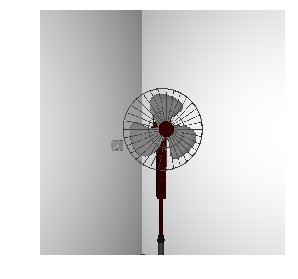}}
&
\subf{\includegraphics[width=\evr]{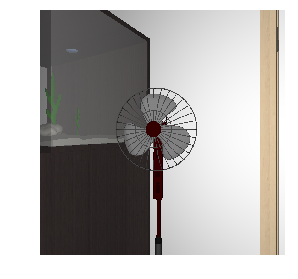}}
&
\subf{\includegraphics[width=\evr]{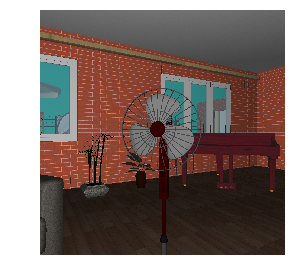}}
&
\subf{\includegraphics[width=\evr]{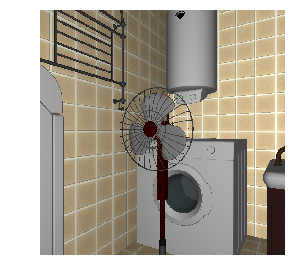}}
\\

\rotatebox{90}{\hspace{30pt}\textbf{piano}}
&
\subf{\includegraphics[width=\evr]{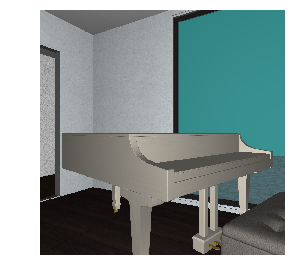}}
&
\subf{\includegraphics[width=\evr]{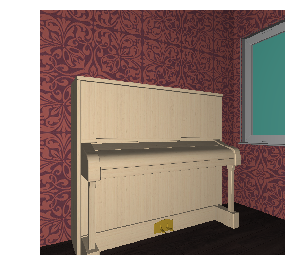}}
&
\subf{\includegraphics[width=\evr]{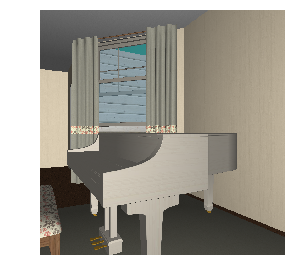}}
&
\subf{\includegraphics[width=\evr]{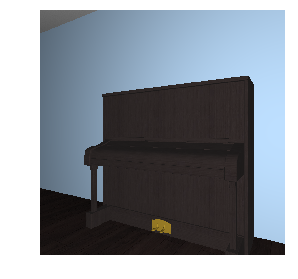}}
\\

\rotatebox{90}{\hspace{20pt}\textbf{fish tank}}
&
\subf{\includegraphics[width=\evr]{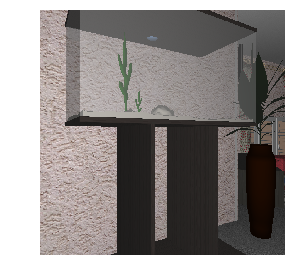}}
&
\subf{\includegraphics[width=\evr]{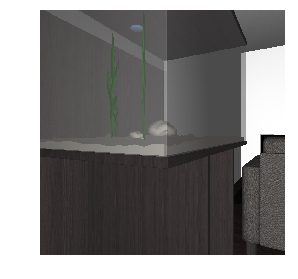}}
&
\subf{\includegraphics[width=\evr]{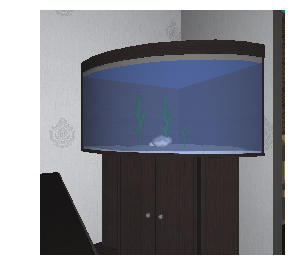}}
&
\subf{\includegraphics[width=\evr]{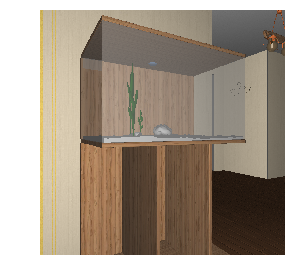}}
\\

\end{tabular}}
\label{fig:suncg-obj}
\caption{Visualizations of queryable objects from the House3D renderer. Notice that instances
within the same class differ significantly in shape, size, and color.}
\end{figure*}

\begin{figure*}
\centering
	\def\evlen{1.0\textwidth}
	\def\evr{0.16\textwidth}
\setlength{\tabcolsep}{0pt}
\resizebox{\textwidth}{!}{    
\begin{tabular}{c c c c c }

\rotatebox{90}{\hspace{15pt}\textbf{living room}}
&
\subf{\includegraphics[width=\evr]{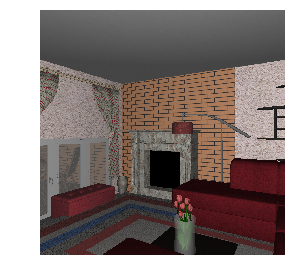}}
&
\subf{\includegraphics[width=\evr]{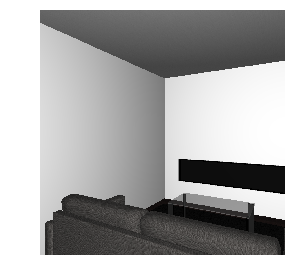}}
&
\subf{\includegraphics[width=\evr]{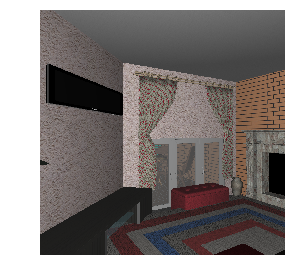}}
&
\subf{\includegraphics[width=\evr]{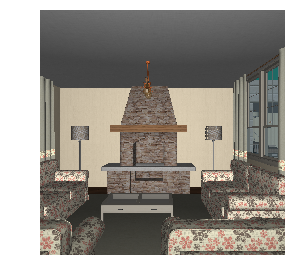}}
\\

\rotatebox{90}{\hspace{25pt}\textbf{kitchen}}
&
\subf{\includegraphics[width=\evr]{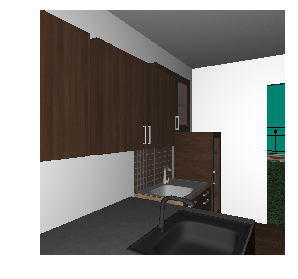}}
&
\subf{\includegraphics[width=\evr]{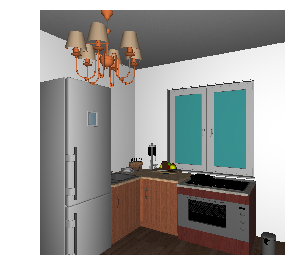}}
&
\subf{\includegraphics[width=\evr]{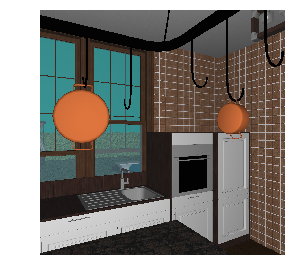}}
&
\subf{\includegraphics[width=\evr]{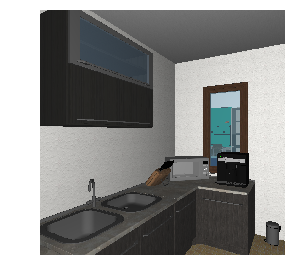}}
\\

\rotatebox{90}{\hspace{20pt}\textbf{bedroom}}
&
\subf{\includegraphics[width=\evr]{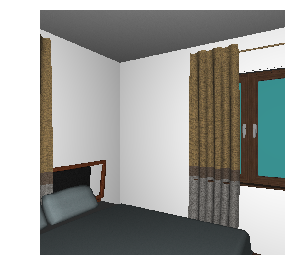}}
&
\subf{\includegraphics[width=\evr]{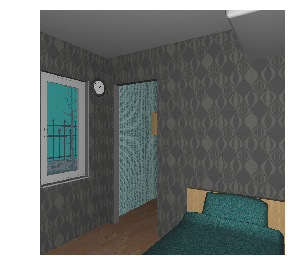}}
&
\subf{\includegraphics[width=\evr]{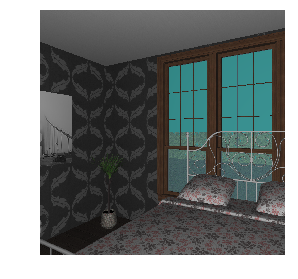}}
&
\subf{\includegraphics[width=\evr]{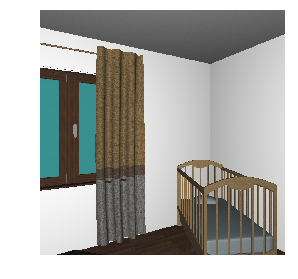}}
\\

\end{tabular}}
\label{fig:suncg-rooms}
\caption{Visualizations of queryable rooms from the House3D renderer.}
\end{figure*}

\begin{figure*}
\centering
\begin{subfigure}[b]{\textwidth}
\centering
\includegraphics[scale=0.24, trim=0cm 25cm 46cm 0cm]{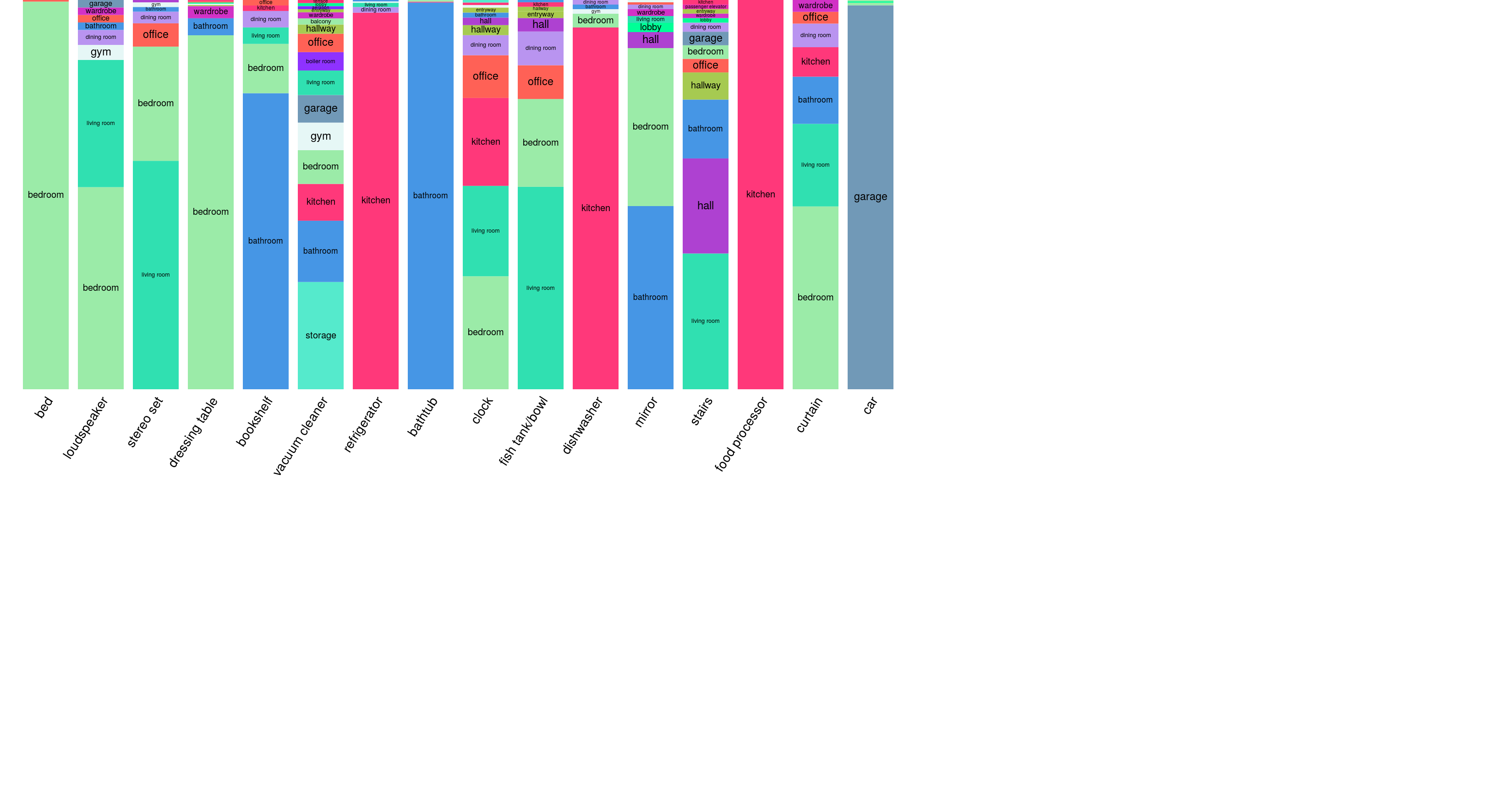}
\caption{\texttt{location} questions before entropy+count based filtering}
\label{fig:loc-before}
\end{subfigure}

\par \bigskip

\begin{subfigure}[b]{\textwidth}
\centering
\includegraphics[scale=0.24, trim=0cm 25cm 46cm 0cm]{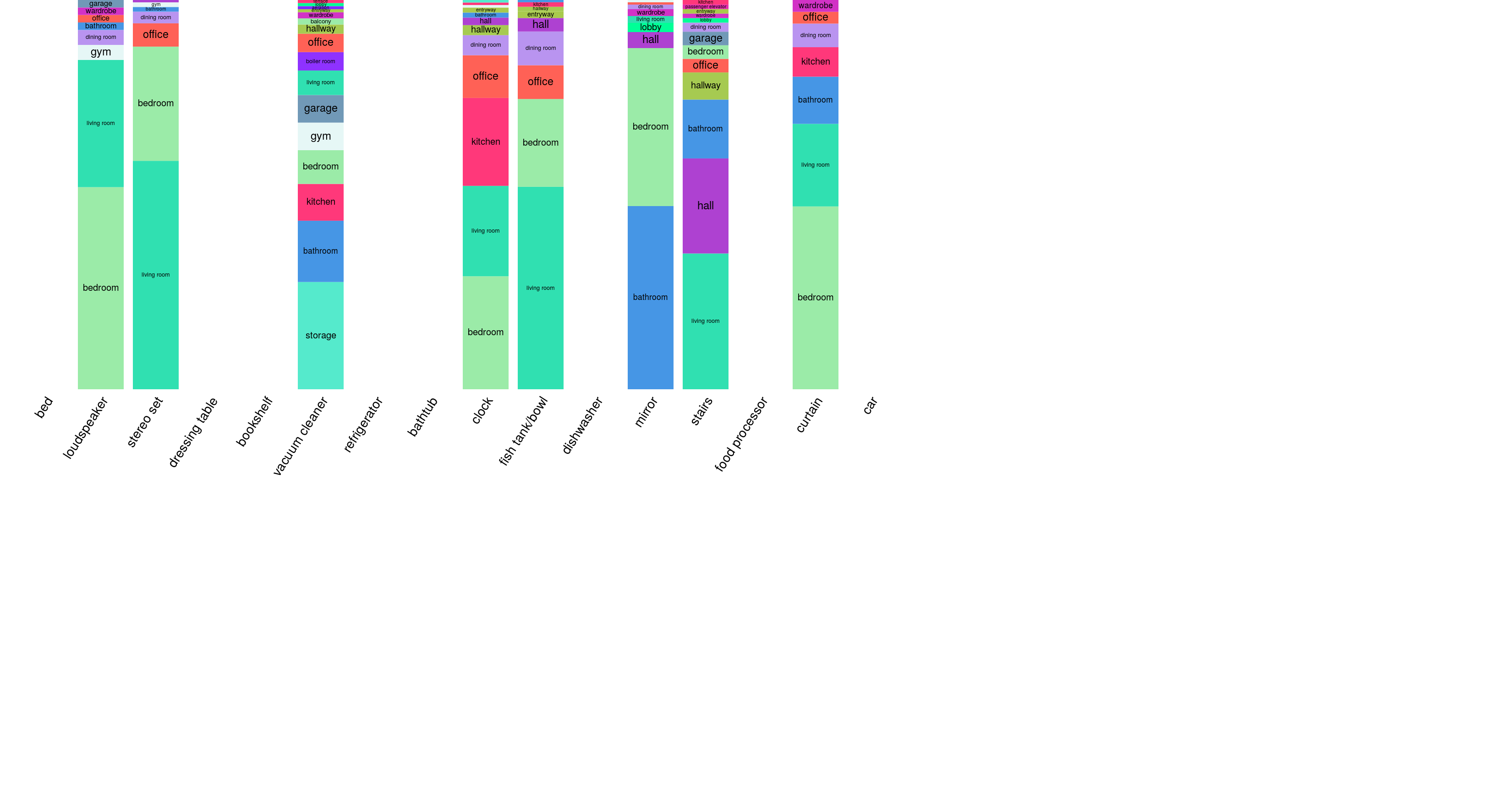}
\caption{\texttt{location} questions after entropy+count based filtering}
\label{fig:loc-after}
\end{subfigure}

\caption{The answer distribution for \texttt{location} template questions.
Each bar represents a question of the form \myquote{what room is the <OBJ> located in?}
and shows a distribution over the answers across different environments. The blank spaces in \ref{fig:loc-after} represent the questions that get pruned out
as a result of the entropy+count based filtering.}
\label{fig:loc-answer-dist}
\end{figure*}

\begin{figure*}
\centering
\begin{subfigure}[b]{\textwidth}
\centering
\includegraphics[scale=0.24, trim=0cm 25cm 54cm 0cm]{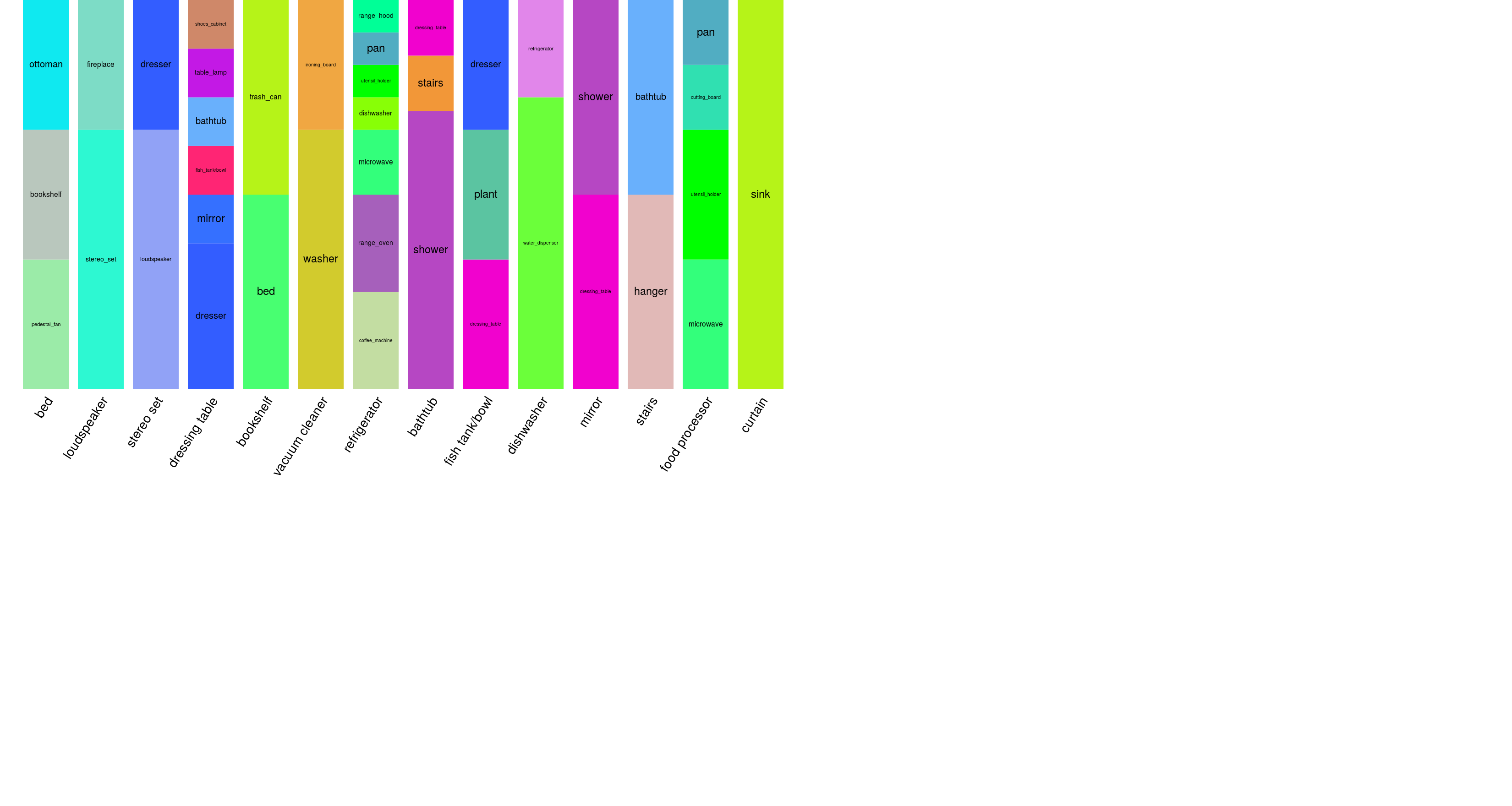}
\caption{\texttt{preposition} questions before entropy+count based filtering}
\label{fig:nextto-before}
\end{subfigure}

\par \bigskip

\begin{subfigure}[b]{\textwidth}
\centering
\includegraphics[scale=0.24, trim=0cm 25cm 54cm 0cm]{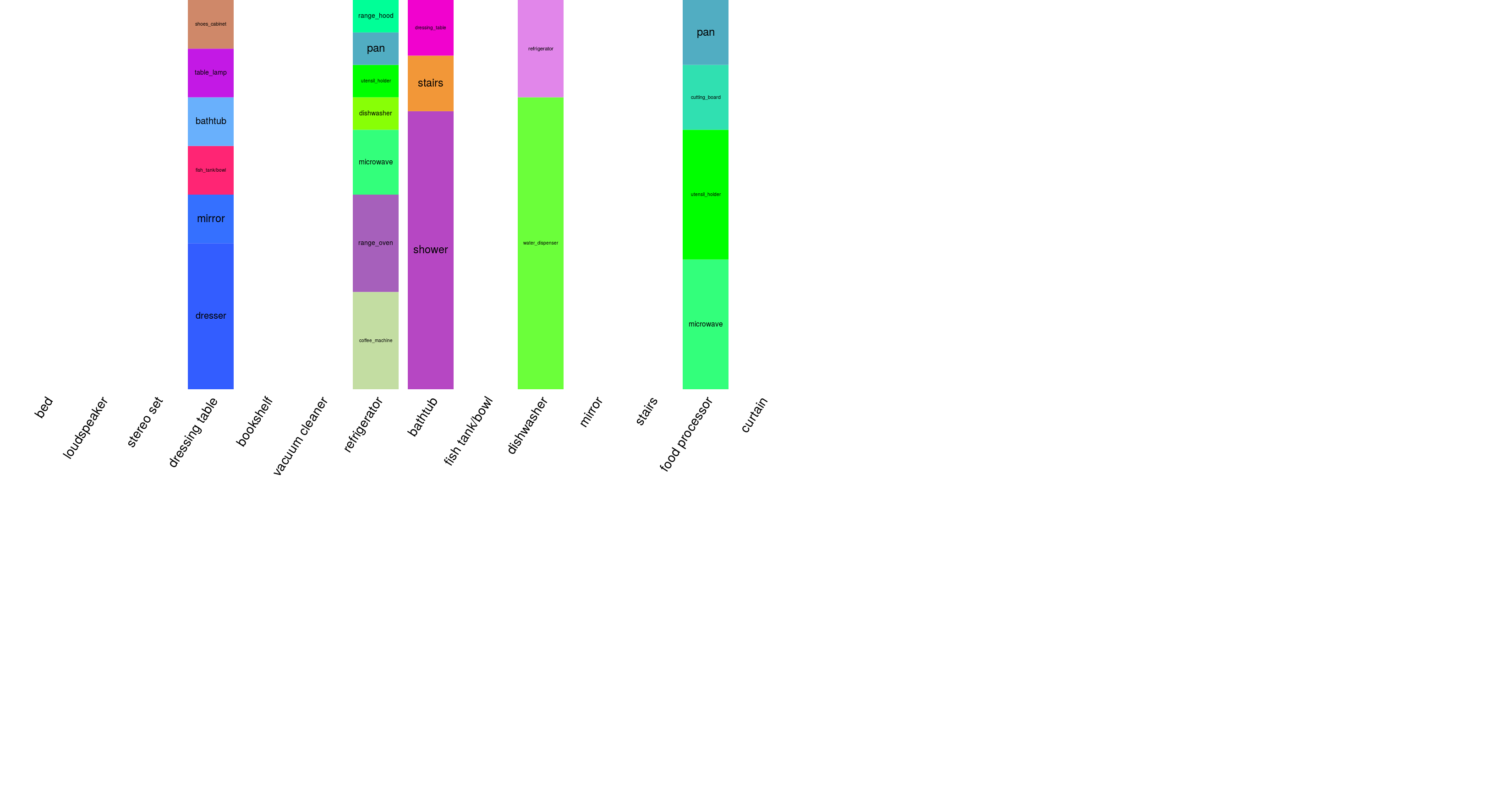}
\caption{\texttt{preposition} questions after entropy+count based filtering}
\label{fig:nextto-after}
\end{subfigure}

\caption{The answer distribution for \texttt{preposition} template questions.
Each bar represents a question of the form \myquote{what is next to the <OBJ>?}
and shows a distribution over the answers across different environments. The blank spaces in \ref{fig:nextto-after} represent the questions that get pruned out as a result of the entropy+count based filtering.}
\label{fig:prep-answer-dist}
\end{figure*}

\begin{figure*}
\centering
\begin{subfigure}[b]{\textwidth}
\centering
\includegraphics[scale=0.24, trim=0cm 25cm 68cm 0cm]{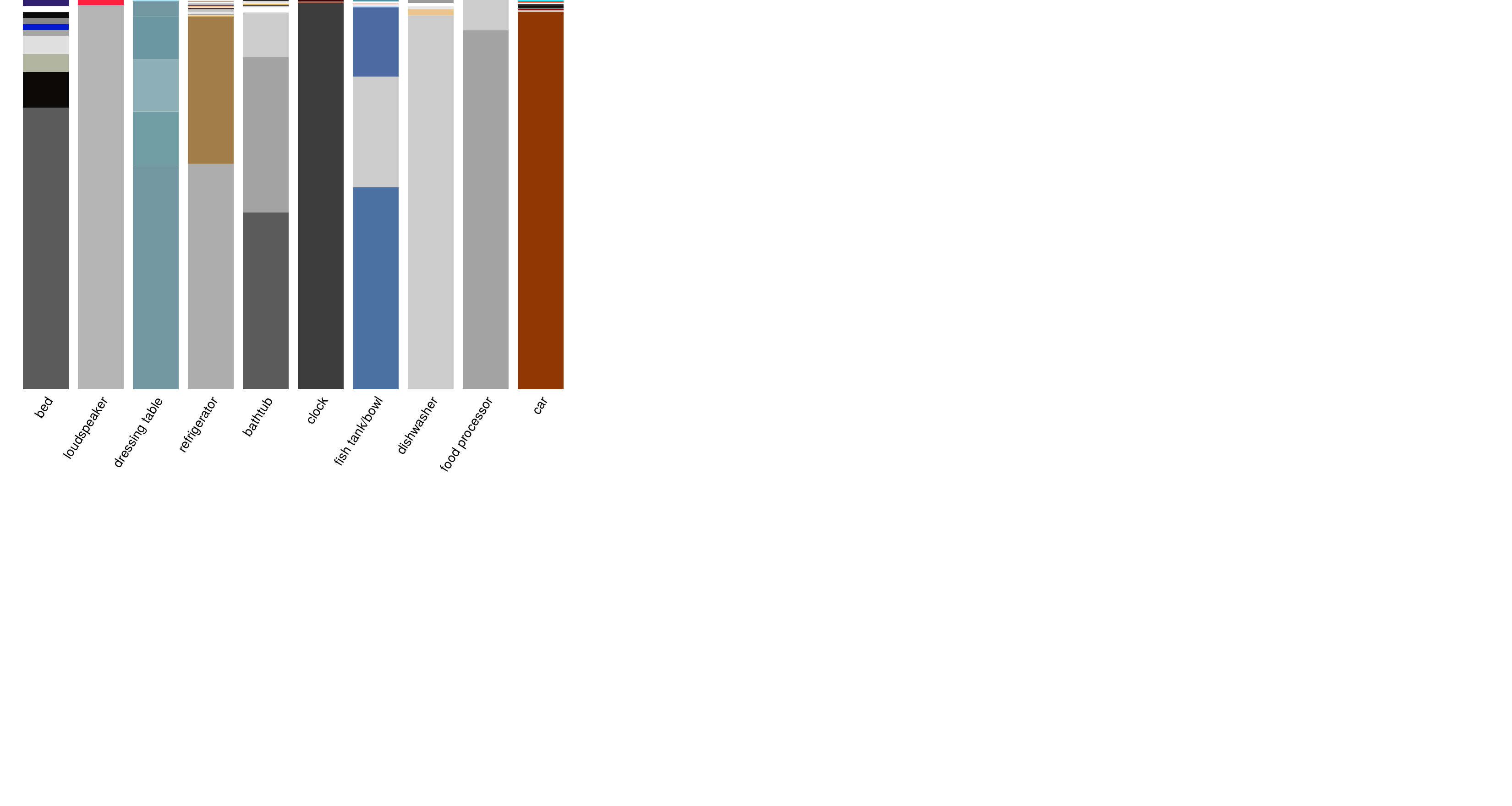}
\caption{\texttt{color} questions before entropy+count based filtering}
\label{fig:color-before}
\end{subfigure}

\par \bigskip

\begin{subfigure}[b]{\textwidth}
\centering
\includegraphics[scale=0.24, trim=0cm 25cm 68cm 0cm]{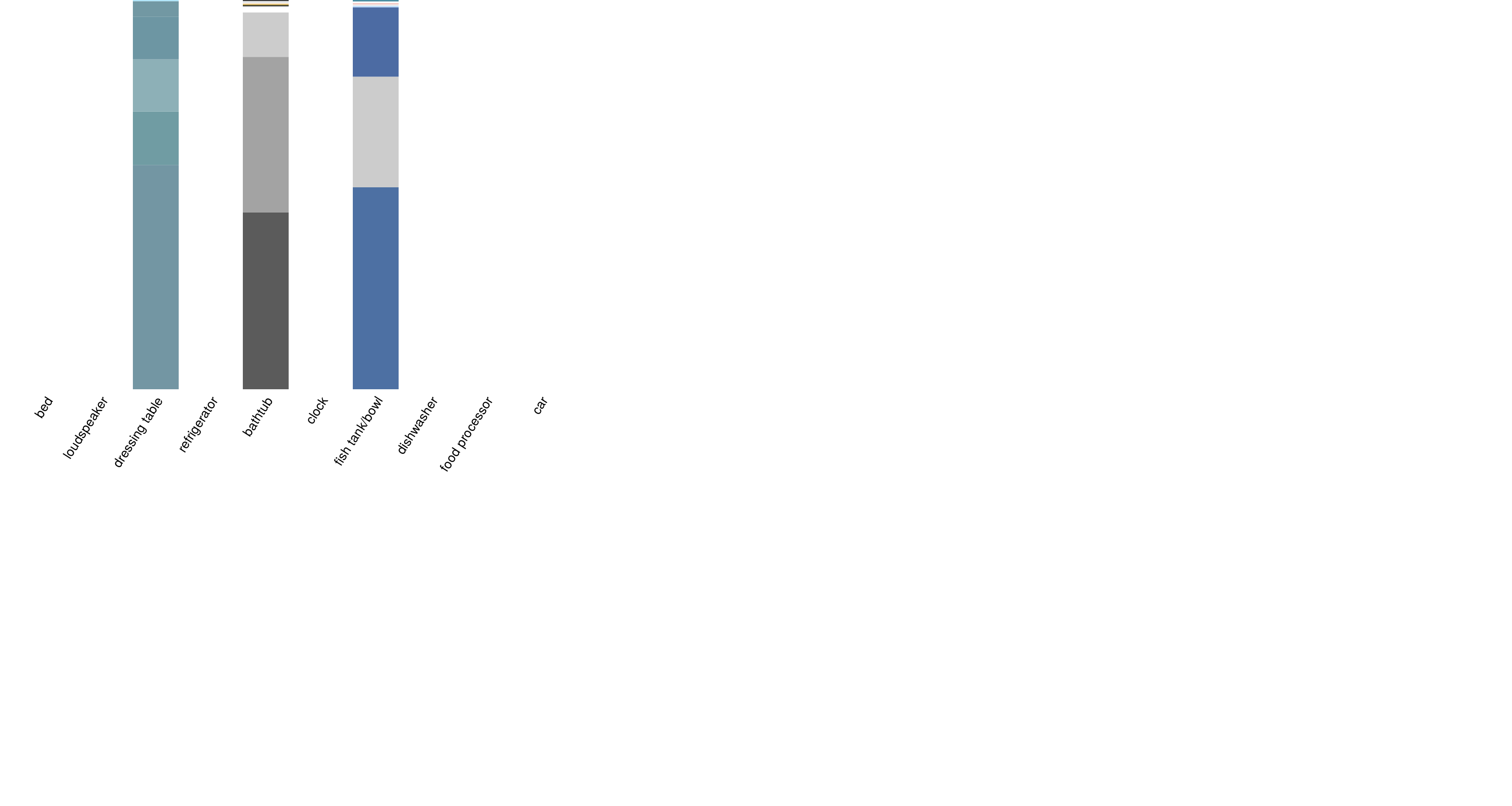}
\caption{\texttt{color} questions after entropy+count based filtering}
\label{fig:color-after}
\end{subfigure}

\caption{The answer distribution for \texttt{color} template questions.
Each bar represents a question of the form \myquote{what color is the <OBJ>?}
and shows a distribution over the answers across different environments
(the color of each section on a bar denotes the possible answers).
The blank spaces in \ref{fig:color-after} represent the questions that get pruned out as a result of the entropy+count based filtering.}
\label{fig:color-answer-dist}
\end{figure*}

\clearpage

{
\small
\bibliographystyle{ieeetr}
\bibliography{strings,main}
}

\end{document}